%% file: main.tex
\documentclass[10pt,twocolumn,letterpaper]{article}

\usepackage[pagenumbers]{cvpr} % To force page numbers, e.g. for an arXiv version


\definecolor{cvprblue}{rgb}{0.21,0.49,0.74}
\usepackage[pagebackref,breaklinks,colorlinks,allcolors=cvprblue]{hyperref}
\usepackage{xcolor} % For \textcolor
\usepackage[table]{xcolor} % For table color
\usepackage{multirow} % For table layout

\usepackage{algorithm}
\usepackage{algpseudocode}
\algtext*{EndIf}
\algtext*{EndFor}
\algtext*{EndFunction}

\usepackage{subcaption}
\usepackage[font=small,labelfont=bf]{caption}
\title{POCA: Pareto-Optimal Curriculum Alignment for Visual Text Generation}

\author{Yaohou Fan$^{1}$, Qingzhong Wang$^2$, Yongsong Huang$^{1}$, Junyi Liu$^2$, Tomo Miyazaki$^{1}$, Shinichiro Omachi$^{1}$\thanks{Corresponding author.}\\
$^1$Tohoku University
$^2$Amazon Web Services\\
{\tt\scriptsize 
\{fan.yaohou.t4\}@dc.tohoku.ac.jp, 
\{hys, tomo, shinichiro.omachi.b5\}@tohoku.ac.jp,
\{qzwang, liujunyi\}@amazon.com
}
}

\begin{document}
\maketitle
\input{sec/0_abstract}    
\input{sec/1_intro}

\input{sec/2_relatedwork}

\input{sec/3_preliminary}
\input{sec/4_method}

\input{sec/5_experiments}
\input{sec/6_conclusion}

{
    \small
    \bibliographystyle{ieeenat_fullname}
    \bibliography{main}
}

\input{sec/X_suppl}

\end{document}

%% file: sec/0_abstract.tex
\begin{abstract}
Current visual text generation models struggle with the trade-off between text accuracy and overall image coherence. We find that achieving high text accuracy can reduce aesthetic quality and instruction-following capability. Although reinforcement learning approaches can alleviate the problem through aligning with multiple rewards, they are often unstable for text generation, as existing approaches normally optimize multiple rewards in a weighted-sum way. In addition, it is difficult to balance the weight of each reward. Moreover, reinforcement learning requires a set of training instructions. A large number of prompts require more training time and computing resources, while a small set leads to poor performance. Hence, how to select the prompts for efficient training is an unsolved problem.
In this study, we propose \underline{P}areto-\underline{O}ptimal \underline{C}urriculum \underline{A}lignment (POCA), a framework that addresses this issue as a multi-objective problem by: 1) identifying the Pareto-optimal set to avoid simple scalarization and 2) designing an adaptive curriculum alignment strategy to manage a learning sequence of a multi-reward dataset using automatic difficulty assessment, which is crucial for optimal convergence as RL methods explore in a limited data environment. In synergy, POCA finds the Pareto-optimal set in a unified reward space, which eliminates inconsistent signals to find the best trade-off solution from different rewards under an easy-to-hard optimization landscape. The experimental results show that POCA significantly improves all metrics such as CLIP, HPS scores and sentence accuracy.

\end{abstract}

%% file: sec/1_intro.tex
\vspace{-1em}
\section{Introduction}
\label{sec:intro}

\begin{figure}[t]
  \centering
  \captionsetup{belowskip=-10pt}
  \includegraphics[width=\linewidth]{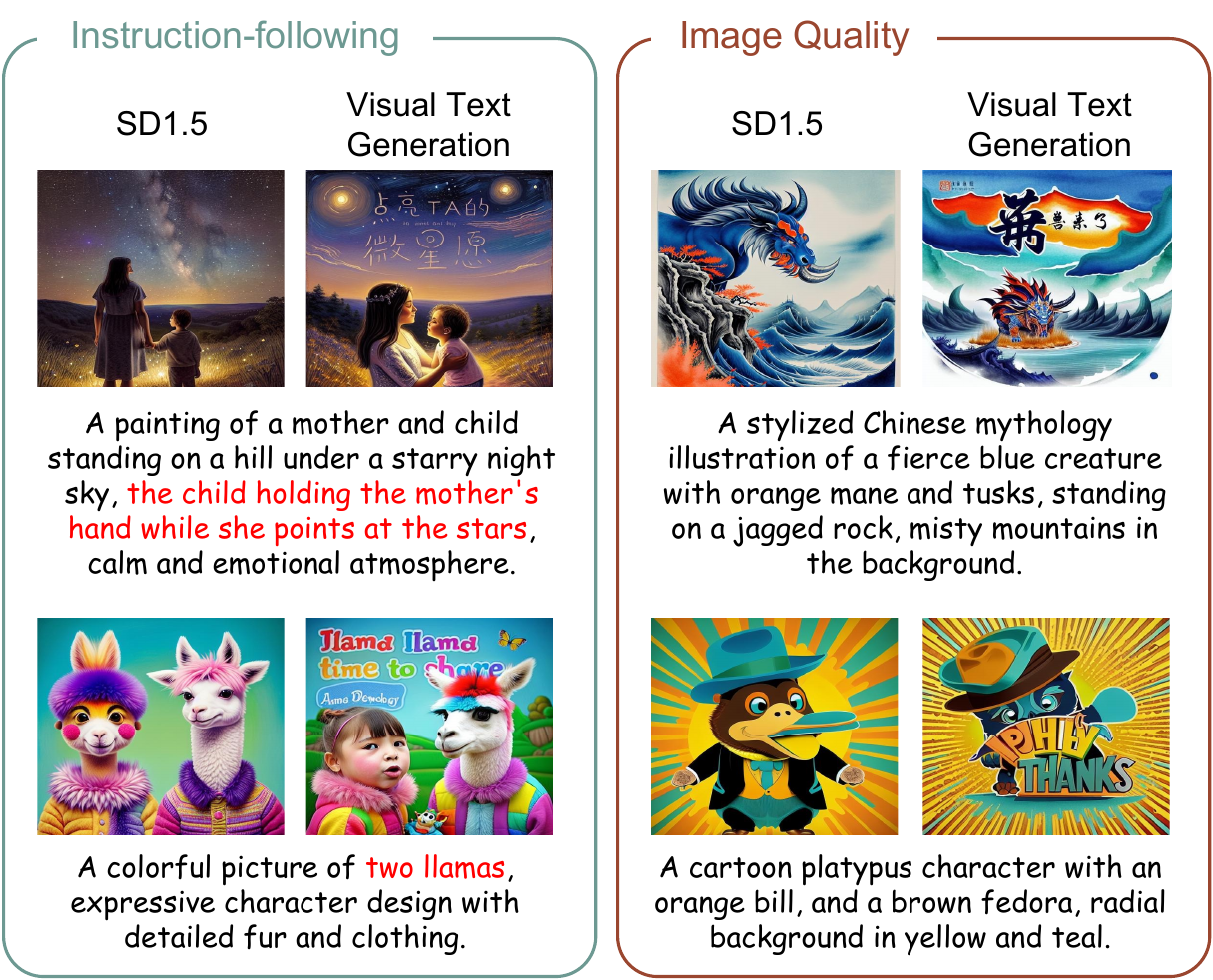}
  \vspace{-1.5em}
  \caption{\textbf{Illustration of the trade-off between image coherence and textual accuracy.} The text generation model shows poor prompt-following and aesthetic appeal ability for generating text.}
  \label{fig:issue}
  \vspace{-1em}
\end{figure}

Advancing image generation capabilities is among the most active research areas in recent years, and visual text generation has emerged as a particularly intensive direction of study. State-of-the-art models, such as Qwen-Image~\cite{qwen} and Seedream~\cite{seedream}, demonstrate a remarkable ability to render plausible text within high-quality background images. However, practical applications frequently require human intervention, such as the precise specification of text regions and fonts for design-like images. This necessity has led to a growing body of work in controllable visual text generation, which typically conditions the generation process on auxiliary signals. These methods achieve more controllable text synthesis by injecting auxiliary information, such as text glyph~\cite{glyphcontrol,anytext,glyphdraw2} and positional cues~\cite{postermaker,udifftext} during generation.

\begin{figure}[t]
  \centering
  \captionsetup{belowskip=-12pt}
  \includegraphics[width=\linewidth]{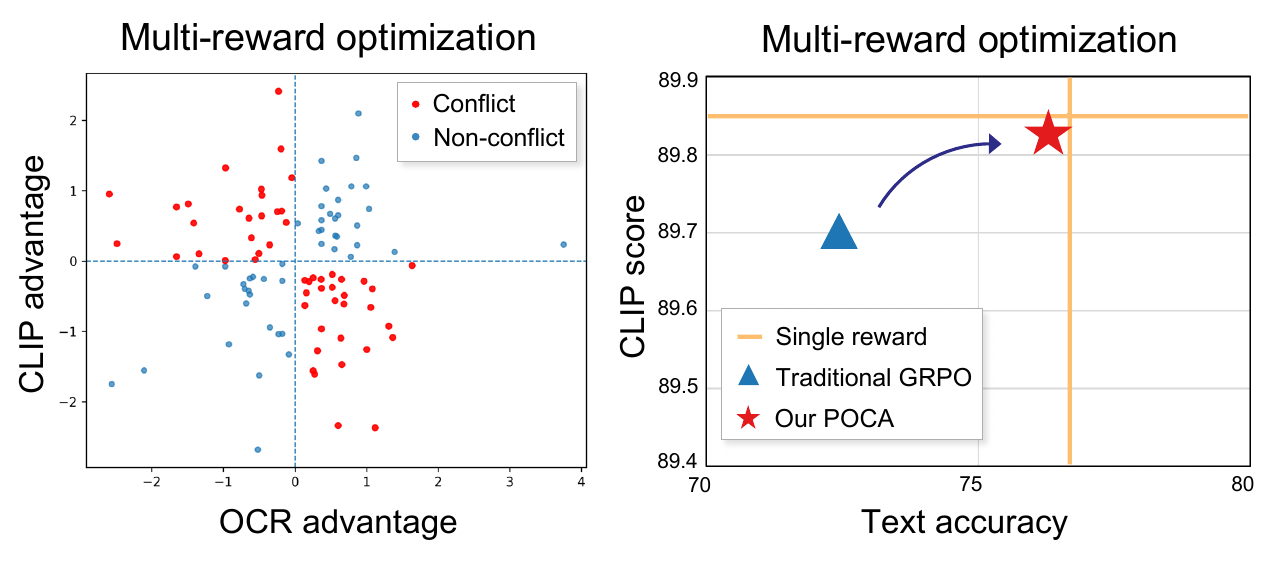}
  \vspace{-2em}
  \caption{\textbf{Inspection of GRPO-based methods in multi-reward optimization.} The figure on the left shows 50 conflicting reward signals among 100 training samples during traditional GRPO training which leads to suboptimal results.}
  \label{fig:motivation}
  \vspace{-0.6em}
\end{figure}

However, following two different instructions during generation introduces an inherent trade-off between the fidelity of the rendered text and the overall image quality. This causes current methods struggle to balance textual accuracy and overall image coherence during training~\cite{glyphdraw2}, as illustrated in Fig.~\ref{fig:issue}. Recently, Reinforcement Learning (RL) has emerged as a tool for adjusting models during post-training to gain proficiency in image generation tasks. Innovative frameworks such as Group Relative Policy Optimization (GRPO)~\cite{dancegrpo,flowgrpo} significantly improve the ability of the base model in a multi-reward optimization task, including prompt alignment and aesthetic criteria. However, they optimize each reward independently and aggregate these separate results using a weighted-sum approach~\cite{dancegrpo,weighted-sum} to update the model. Our empirical analysis reveals that this weighted-sum approach destabilizes the GRPO training process due to contradictory reward signals (e.g., OCR and CLIP) in the visual text generation task, as shown in Fig.~\ref{fig:motivation}. This simple aggregation results in an ambiguous optimization target, leading to suboptimal convergence.

Moreover, prompt selection for RL is an unsolved problem. Using static datasets for multi-reward optimization introduces two drawbacks. First, although performance often improves with data quantity, simply scaling to very large datasets is computationally expensive and requires more training time. Second, multi-reward optimization requires a diverse mixture of prompts from different domains to provide targeted signals for each reward objective, for example, mixing prompts that are biased towards aesthetic descriptions with others biased towards textual accuracy. This might yield a complex optimization landscape that slows convergence. This raises a practical question: how should we select prompts to train both efficiently and effectively? 

In this study, we address the aforementioned issues by introducing POCA (Pareto-Optimal Curriculum Alignment), a synergistic framework composed of two core components. First, to overcome the limitations of the weighted-sum approach, our Pareto-optimal selection algorithm reframes the task as a multi-reward optimization problem. Instead of collapsing rewards, it identifies the Pareto set in the joint reward space. The idea is drawn from Pareto-set learning~\cite{pareto_define,paretoprinciple}, which states that the solution to a conflicting multi-objective optimization problem is the best trade-off state among objectives, referred to as Pareto-optimality. Therefore, we propose a bi-directional sorting algorithm that finds optimal positive and negative samples using all rewards simultaneously and only uses these samples for optimization. This selection removes inconsistent signals during GRPO training. 

Second, from the data perspective, we propose a curriculum planning strategy that effectively selects and manages training data to expose models to increasingly complex environments that mimic the human learning path. An adaptive strategy is proposed to ensure a steady progression of model capabilities through a difficulty-aware design. Specifically, we dynamically compile the Empirical Cumulative Distribution Function (ECDF) of rewards for difficulty assessment of a multi-reward dataset before and during training, so that it is tightly coupled to model growth. Based on this assessment, we are able to adaptively select optimal curricula to learn the best trade-off state under a complex reward environment. Our contributions can be summarized as follows:
\begin{itemize}
    \item We find that GRPO-based methods become unstable for conflicting rewards. Therefore, we introduce a bi-directional Pareto sorting that retains only the valid reward signals for optimization, which stabilizes the training.
    
    \item We design an adaptive curriculum planning strategy for data selection that organizes training data into a dynamic "easy-to-hard" path, achieving both faster convergence and improved final performance.
    
    \item Combining the first two components, we propose POCA, which jointly aligns policy updates with Pareto optimal selection and adapts the learning sequence to task difficulty. This synergy significantly improves the performance of the visual text generation model in text accuracy, text–image alignment, and aesthetic quality.
\end{itemize}

%-------------------------------------------------------------------------

%% file: sec/2_relatedwork.tex
\begin{figure*}[t]
  \centering
  \includegraphics[height=0.55\textwidth,width=0.75\textwidth]{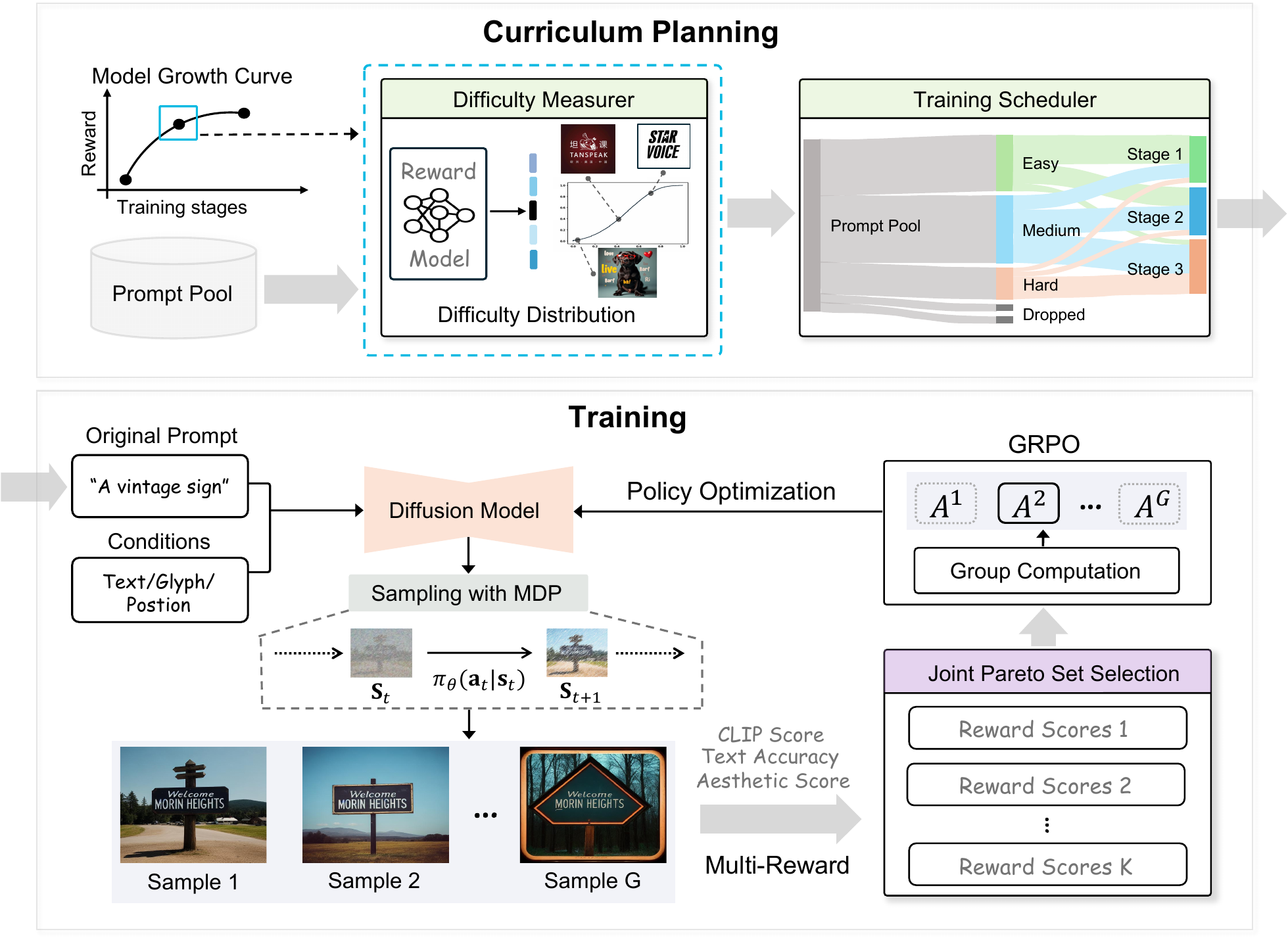}
  \vspace{-0.5em}
  \caption{\textbf{Overview of POCA.} Our framework operates in two stages. The \textbf{Curriculum Planning stage} assesses prompt difficulty and organizes the data into an "easy-to-hard" sequence. This sequence is then fed to the \textbf{Training stage}, which uses bi-directional Pareto sorting to filter for the Pareto set for optimization.}
  \label{fig:workflow}
  \vspace{-0.6em}
\end{figure*}

\section{Related Work}
\label{sec:relatedwork}

\textbf{Visual Text Generation: }
Generating images with high-fidelity text has long been a challenging task. The visual generation community has primarily adopted two mechanisms for injecting auxiliary information during synthesis to enhance this capability: (1) prompt-based conditioning and (2) model-based conditioning. On the one hand, early work demonstrated that character-aware text encoders~\cite{byt5} significantly improve the text-rendering accuracy of diffusion models compared to their character-blind counterparts~\cite{clip,t5}. Following this insight, a series of works~\cite{udifftext,diffste,textdiffuser2,glyphbyt5} adopted various character-level encoders to extract more robust and detailed text features as conditional guidance. However, prompt-based conditioning does not fully meet all user needs. Hence, GlyphControl~\cite{glyphcontrol} proposed model-based conditioning by introducing ControlNet for text rendering under specific user requirements. AnyText~\cite{anytext}, Brush Your Text~\cite{brush} and FLUX-Text~\cite{fluxtext} further incorporated auxiliary text conditions such as glyph, position and color for controlled text synthesis.

\textbf{Reinforcement Learning from Human Feedback: } 
Reinforcement Learning (RL) methodologies cast a diffusion model as a policy model in a Markov Decision Process (MDP), where the model attempts multiple generation trajectories for a given prompt. Policy optimization methods are then used to up-weight the policy that derived the trajectory with higher reward. For reward models, a number of works have been proposed, which are trained on massive human-annotated preference data to supply dense preference signals that guide policy explorations~\cite{imagereward,hpsscore,pickscore}. For policy optimization, traditional methods~\cite{rwr1,rwr2,wang2019describing,wang2020diversity} simply re-weight denoising loss with terminal rewards. This is proven to be only a rough approximation of the solution. Direct Policy Optimization (DPO) foregoes explicit reward models. Instead, it uses the exact same pairwise preference data as human feedback rewards to assign credit across timesteps~\cite{dpo,ddpo}. GRPO~\cite{deepseek,dancegrpo,flowgrpo,mixgrpo} is introduced to evaluate trajectories in a group-relative manner which is more stable and scalable than DPO. It uses the average reward of multiple sampled outputs, produced from the same conditional prompt, as the baseline and optimizes the base model according to the resulting intra-group ranking. 

\textbf{Curriculum-driven Learning: }
In human learning domains, especially for hard tasks such as math learning, the quality of a curriculum has been shown to be crucial in achieving success. In an RL setting, CL aims to optimize the order in which experience is accumulated by the policy model, so as to increase performance or training speed on a set of final tasks. In the most general case, a Difficulty Measurer and a Training Scheduler can be used to design a learning sequence as an optimal curriculum by sorting and managing training samples~\cite{cl_survey}. Recent works employ difficulty-based sorting algorithms~\cite{difficultymeasurer1,difficultymeasurer2,curriculumDPO} to produce a difficulty ranking of data. The training scheduler turns the ranking into a predetermined curriculum progression~\cite{boostdpo,curricula}. In contrast to these predefined, human-driven CL strategies, we design a model-driven CL method with an automatic difficulty measurer and training scheduler to adaptively adjust training samples in a self-paced manner, which is similar to active learning~\cite{zhanpareto,zhan2022comparative}.

%% file: sec/3_preliminary.tex
\section{Preliminary}

In this section, we briefly describe how GRPO-based RL algorithms train diffusion models by formulating the denoising process as an MDP.

\textbf{Denoising in MDP framework: } As shown in~\cite{ddpo}, the iterative denoising procedure of diffusion model can be mapped to a multi-step MDP $(\mathcal{S}, \mathcal{A}, \rho_0, P, R)$.
The state at step $t$ is $\mathbf{s}_t \triangleq (\mathbf{c}, t, \mathbf{z}_t)$, where $\mathbf{c}$ is the prompt. The action is the denoised sample $\mathbf{a}_t \triangleq \mathbf{z}_{t-1}$ predicted by the model, and the policy is $\pi(\mathbf{a}_t \mid \mathbf{s}_t) \triangleq p_\theta(\mathbf{z}_{t-1} \mid \mathbf{z}_t, \mathbf{c})$. The reward is only given at the final step: $R(\mathbf{s}_t, \mathbf{a}_t) \triangleq r(\mathbf{z}_0, \mathbf{c})$ if $t=0$, and $0$ otherwise. As the diffusion model acts according to the policy $\pi(\mathbf{a} \mid \mathbf{s})$, it produces trajectories contain sequences of states and actions $\tau = (\mathbf{s_0}, \mathbf{a_0}, \mathbf{s_1}, \mathbf{a_1}, \dots, \mathbf{s_T}, \mathbf{a_T})$. The objective of standard RL policy gradient method is to maximize the expected reward of trajectories:
\begin{equation}
\mathcal{J}_{\mathrm{RL}}(\pi)
= \mathbb{E}_{\tau \sim p(\tau \mid \pi)}
\!\left[ \sum_{t=0}^{T} R(\mathbf{s}_t, \mathbf{a}_t) \right].
\end{equation}

\textbf{Group Relative Policy Optimization: } GRPO~\cite{deepseek} modifies the standard policy gradient by introducing relative advantages within sets of outputs corresponding to the same query for stabilizing updates. Given a prompt $\mathbf{c}$, generative models will sample a group of outputs $\{ \mathbf{o}_1, \mathbf{o}_2, ..., \mathbf{o}_G \}$ from the model $\pi_{\theta_{old}}$, and the corresponding rewards are $\{ r_1, r_2,...,r_G \}$. GRPO optimizes the policy model $\pi_{\theta}$ by maximizing the following objective function:
\begin{multline}
\mathcal{J}(\theta)
= \mathbb{E}_{\substack{\{\mathbf{o}_i\}_{i=1}^G \sim \pi_{\theta_{\text{old}}}(\cdot|\mathbf{c}) \\ \mathbf{a}_{t,i} \sim \pi_{\theta_{\text{old}}}(\cdot|\mathbf{s}_{t,i})}}
\Bigg[ \frac{1}{G} \sum_{i=1}^G \frac{1}{T} \sum_{t=1}^T \\
\min\Big( \mathbf{\rho}_{t,i} A_i,\; \mathrm{clip}\big(\mathbf{\rho}_{t,i},1-\epsilon,1+\epsilon\big)\,A_i \Big) \Bigg],
\label{eq:dancegrpoloss}
\end{multline}
where $\mathbf{\rho}_{t,i} = \frac{\pi_{\theta}(\mathbf{a}_{t,i}|\mathbf{s}_{t,i})}{\pi_{\theta_{old}}(\mathbf{a}_{t,i}|\mathbf{s}_{t,i})} $, and $\pi_{\theta}(\mathbf{a}_{t,i}|\mathbf{s}_{t,i})$ is the policy function for output $\mathbf{o}_i$ at time step $t$, $\epsilon$ is a hyper-parameter, and $A_i$ is the advantage function defined by:
\begin{equation}
    A_i=\frac{r_i-\mathrm{mean}(\{r_1,r_2,\cdots,r_G\})}{\mathrm{std}(\{r_1,r_2,\cdots,r_G\})}.
\label{eq:adv}
\end{equation}

\iffalse
\begin{equation}
\mathcal{J}_{\mathrm{RL}}(\pi)
= \mathbb{E}_{\tau \sim p(\tau \mid \pi)}
\!\left[ \sum_{t=0}^{T} R(\mathbf{s}_t, \mathbf{a}_t) \right].
\end{equation}
\fi

%% file: sec/4_method.tex
\section{Methodology}

\begin{algorithm*}[t]
\small
\caption{Pareto-guided GRPO with Bi-directional Pareto Sorting}
\label{algo:pocagrpo}

\begin{minipage}{0.50\textwidth}
\footnotesize
\begin{algorithmic}
\Require Initial policy model $\pi_\theta$; reward models $\{R_k\}_{k=1}^K$; 
         prompt dataset $\mathcal{D}$; total sampling steps $T$;
\Ensure Pareto-guided optimization

\For{training step $=1$ \textbf{to} $M$}
    \State Sample batch $\mathcal{D}_b \sim \mathcal{D}$ 
    \State Update old policy: $\pi_{\theta_{\text{old}}} \gets \pi_\theta$
    
    \For{each prompt $\mathbf{c} \in \mathcal{D}_b$}
        \State Generate $G$ samples: $\{\mathbf{o}_i\}_{i=1}^G \sim \pi_{\theta_{\text{old}}}(\cdot|\mathbf{c})$
        \State Compute rewards $\{r_i^k\}_{i=1}^G$ using each $R_k$
        \State $\mathcal{P} \gets \Call{BiNDSet}{\{r_i^k\}_{i=1}^G}$
    
        \For{each sample $i \in 1..G$}
            \State Calculate multi-reward advantage: 
            $A_i$
        \EndFor
        
        \For{$t \in T$}
            \State Update policy via gradient ascent:
            $\theta \gets \theta + \eta \nabla_\theta \mathcal{J}_\mathcal{P}$
        \EndFor
    \EndFor
\EndFor
\end{algorithmic}
\end{minipage}\hfill
\begin{minipage}{0.45\textwidth}
\footnotesize
\begin{algorithmic}
\Function{NDSet}{$\{r_i^k\}_{i=1}^G$} 
    \State $\mathcal P \gets \emptyset$
    \For{$i=1$ \textbf{to} $G$}
        \State ${dominated} \gets \textbf{False}$
        \For{$j=1$ \textbf{to} $G$}
            \If{$(\forall k:\;r_j^{(k)}\!\ge r_i^{(k)})\ \wedge\ (\exists k:\;r_j^{(k)}\!>\! r_i^{(k)})$}
                \State ${dominated}\gets \textbf{True}$ 
            \EndIf
        \EndFor
        \If{\textbf{not} $dominated$} \State Add $i$ to $\mathcal P$ \EndIf
    \EndFor
    \State \Return $\mathcal P$
\EndFunction

\Function{BiNDSet}{$\{r_i^k\}_{i=1}^G$}
    \State $\mathcal P^+ \gets \Call{NDSet}{\{r_i^k\}_{i=1}^G}$ \Comment{Positive Pareto set}
    \State $\mathcal P^- \gets \Call{NDSet}{\{-r_i^k\}_{i=1}^G}$ \Comment{Negative Pareto set}
    \State \Return $(\mathcal P^+,\mathcal P^-)$
\EndFunction
\end{algorithmic}
\end{minipage}

\end{algorithm*}

%\subsection{POCA Overview}

In this section, we describe the POCA workflow. POCA applies GRPO-based reinforcement learning method to enhance current text generation models with multiple rewards, including prompt alignment, aesthetic quality and text accuracy. Fig.~\ref{fig:workflow} shows POCA consists of a Pareto-optimal selection and a curriculum planning stage compared with traditional GRPO. 

In the post-training stage, given the same original prompt and corresponding text conditions, we sample a group of outputs and calculate the aforementioned rewards for each image, yielding a set of reward scores. Based on these scores, POCA employs a bi-directional sorting algorithm to identify the Pareto set within the joint reward space. Finally, only this optimal set is used to update the model parameters via policy gradient ascent.

Moreover, this entire optimization process is guided by our adaptive curriculum strategy. It is designed to periodically assess current model capabilities using reward models against the full data environment. Based on this difficulty assessment, it adaptively schedules a learning sequence, managing which subset of prompts is fed into the training stage for the model to explore. This ensures the model learns from an "easy-to-hard" optimization landscape, mastering simpler trade-offs before progressing to more complex ones.

\subsection{Joint Pareto Set Selection}

As established in the introduction, the standard GRPO framework is unstable for balancing conflicting objectives, especially in visual text generation. Therefore, we seek a solution that balances all rewards well to solve this multi-objective problem. Pareto-set learning has demonstrated its effectiveness in optimizing conflicting objectives simultaneously~\cite{pareto,parrot}. It aims to achieve the best trade-off state for multiple objectives where one objective cannot be improved without another worsening. Applied to our multi-reward setting, each generated text image presents a particular trade-off among multiple rewards, and some samples present an optimal trade-off that is both textually accurate and aesthetically pleasing. We refer to such samples as Pareto-optimal (non-dominated) samples, and our model should be optimized in a way that generates more Pareto-optimal samples. POCA achieves this by identifying: 1) the non-dominated set and 2) the fully dominated set in a unified reward space via a bi-directional Pareto sorting algorithm for policy updates. Algorithm~\ref{algo:pocagrpo} outlines this procedure. In practice, the non-dominated set provides positive signals, and the fully dominated set provides negative signals, which GRPO (Eq.~\ref{eq:dancegrpoloss}) uses to increase and decrease, respectively, their likelihood of being generated during training.

\textbf{Bi-directional Pareto Sorting:} Within each GRPO group of $G$ sampled outputs, we perform a group-based Pareto sorting to obtain a Pareto set $\mathcal{P}$ containing the non-dominated set and the fully dominated set. Let \(\mathcal{O}=\{\mathbf{o}_1,\ldots,\mathbf{o}_G\}\) be a group of sampled outputs and \(R(\mathbf{o})=(R_1(\mathbf{o}),\ldots,R_K(\mathbf{o}))\) the \(K\)-dimensional reward vector. The dominance relationship can be described as: \(\mathbf{o}_1\) dominates \(\mathbf{o}_2\) (denoted \(\mathbf{o}_1 \succeq \mathbf{o}_2\)) if and only if \(R_i(\mathbf{o}_1)\ge R_i(\mathbf{o}_2)\) for all \(i\in\{1,\ldots,K\}\) and there exists \(j\in\{1,\ldots,K\}\) such that \(R_j(\mathbf{o}_1)>R_j(\mathbf{o}_2)\). A sample \(\mathbf{o}_i\) is non-dominated if no point in \(\mathcal{O}\) dominates it. We further identify the fully dominated set by applying non-dominated sorting to the negated rewards, which gives a subset of clearly poor trade-offs for negative credit assignment.

Compared with \cite{parrot}, the proposed bi-directional Pareto sorting approach considers the best and worst samples, where the best samples provide positive advantages and the worst ones provide negative advantages in GRPO. Hence, the model learns to be closer to the best but to keep away from the worst.

\textbf{Pareto-guided GRPO:} We eliminate samples not included in our Pareto set $\mathcal{P}$ by assigning a zero advantage. As a result, only the data points in $\mathcal{P}$ contribute to the objective function as follows:

\begin{multline}
\mathcal{J}_{\mathcal{P}}(\theta)
= \mathbb{E}_{\substack{\{\mathbf{o}_i\}_{i=1}^G \sim \pi_{\theta_{\text{old}}}(\cdot|\mathbf{c}) \\ \mathbf{a}_{t,i} \sim \pi_{\theta_{\text{old}}}(\cdot|\mathbf{s}_{t,i})}}
\Bigg[ \frac{1}{n(\mathcal{P})} \sum_{i \in \mathcal{P}} \frac{1}{T} \sum_{t=1}^T \\
\min\Big( \mathbf{\rho}_{t,i} A_i,\; \mathrm{clip}\big(\mathbf{\rho}_{t,i},1-\epsilon,1+\epsilon\big)\,A_i \Big) \Bigg].
\label{eq:paretogrpo}
\end{multline}

\subsection{Adaptive Curriculum Planning}

Rather than training on a large static dataset, which is computationally expensive, POCA employs an adaptive curriculum planning strategy. It aims to shape a stable learning path and help the policy model to learn robust reward trade-offs from an easy-to-hard data environment. It contains two components: 1) a Difficulty Measurer decides the relative easiness of each prompt, and 2) a Training Scheduler decides the sequence of data subsets throughout the training process based on the judgment from the Difficulty Measurer. 

\textbf{Difficulty Measurer:} During training, the policy model samples outputs for a given prompt, and the reward models score these outputs. We find this procedure naturally assigns a difficulty signal to each prompt. We therefore treat the OCR reward model among all reward models as a Difficulty Measurer to assign difficulty scores for all prompts in the dataset. This choice is motivated by our empirical observation that the distribution of OCR rewards exhibits larger variance across the dataset compared to other rewards as shown in Appendix~\ref{sec:variance_compare}. In practice, for every prompt $\mathbf{c} \in \mathcal{D}$, we first estimate its mean OCR reward by sampling $N$ times with the current policy model $\pi_{\theta}$:
\begin{equation}
\mu_{c} = \frac{1}{N}\sum_{j=1}^{N} R_{\text{ocr}}(\mathbf{o}_{c,j}),
\label{eq:prompt-mean-reward}
\end{equation}
we then normalize these scores using the Empirical Cumulative Distribution Function (ECDF) to obtain a difficulty ranking. Specifically, we apply the ECDF to all mean OCR rewards of the entire dataset $\mathcal{D}$. Formally, the ECDF value $ECDF(x)$ of a prompt (where $x$ is its mean OCR reward $\mu_c$) is the fraction of prompts in the dataset whose mean OCR reward is less than or equal to $x$:
\begin{equation}
\label{eq:ecdf}
ECDF(x) = \frac{1}{|\mathcal{D}|} \sum_{\mathbf{c}' \in \mathcal{D}} \mathbf{1}(\mu_{c'} \leq x),
\end{equation}
here, $\mathbf{1}(\cdot)$ is the indicator function, which evaluates to 1 if the condition is true and 0 otherwise. This resulting ECDF value $ECDF(x) \in [0, 1]$, which represents the percentile rank of a prompt, serves as the normalized relative difficulty ranking. This rank is then passed to the Training Scheduler. Finally, to ensure the curriculum remains tightly coupled with the model's growth, we divide the total training process into $S$ equal stages, and at the beginning of each stage, we re-invoke this entire procedure with current policy model.

\textbf{Training Scheduler:} The Training Scheduler is responsible for converting the difficulty rankings from the Difficulty Measurer into an easy-to-hard training plan. First, at the beginning of each training stage, we trim a small fraction of the easiest and hardest samples, as the model either trivially solves them or fails to learn from them~\cite{RLobjective}. Then, Training Scheduler uses the ECDF value of each prompt to partition the entire dataset $\mathcal{D}$ into three discrete difficulty bins by defining fixed percentile thresholds:
\begin{equation}
\label{eq:difficulty-bins}
\begin{split}
    \mathcal{D}_{easy}   &= \{ \mathbf{c} \in \mathcal{D} \mid ECDF(\mu_c) \geq 0.7 \} \\
    \mathcal{D}_{medium} &= \{ \mathbf{c} \in \mathcal{D} \mid 0.3 < ECDF(\mu_c) < 0.7 \} \\
    \mathcal{D}_{hard}   &= \{ \mathbf{c} \in \mathcal{D} \mid ECDF(\mu_c) \leq 0.3 \}
\end{split}
\end{equation}
after this, the Training Scheduler implements the curriculum by assigning a different mixture ratio $\mathbf{w}_s$ for predefined stage $S$. This bucketing curriculum, outlined in Fig.~\ref{fig:workflow}, progressively exposes the model to more difficult environment. In our default configuration, we set $S=3$ and use $\mathbf{w}_1=$~(0.6, 0.3, 0.1), $\mathbf{w}_2=$~(0.4, 0.5, 0.1), $\mathbf{w}_3=$~(0.1, 0.6, 0.3) for the easy, medium and hard prompt ratios, respectively. It is obviously challenging to find the most suitable combination of these hyper-parameters. This empirical configuration is driven by recent findings that moderate level of difficulty environment is more beneficial for model growth~\cite{limr}.

%% file: sec/5_experiments.tex
\begin{table*}[t]
\caption{\small Quantitative comparison of POCA and visual text generation methods on the AnyText-benchmark. \textcolor{green}{\dag} is our reproduced results based on our implementation using official code.}
\vspace{-6pt}
\label{table:performance_comparison}
\centering
\footnotesize
\resizebox{0.8\textwidth}{!}{%
\begin{tabular}{@{}lcccc|cccc@{}}
\toprule
\multirow{2}{*}{Methods} &
\multicolumn{4}{c|}{English} &
\multicolumn{4}{c}{Chinese} \\
\cmidrule(l){2-5} \cmidrule(l){6-9}
 & Sen.ACC$\uparrow$ & NED$\uparrow$ & CLIP score$\uparrow$ & HPS score$\uparrow$
 & Sen.ACC$\uparrow$ & NED$\uparrow$ & CLIP score$\uparrow$ & HPS score$\uparrow$ \\
\midrule
ControlNet\cite{controlnet} & 0.5837 & 0.8015 & 0.8448 & 0.2257 & 0.3620 & 0.6227 & 0.7801 & 0.2277 \\
TextDiffuser\cite{textdiffuser} & 0.5921 & 0.7951 & 0.8685 & 0.2418 & 0.0605 & 0.1262 & 0.7675 & 0.2335 \\
GlyphControl\cite{glyphcontrol} & 0.5262 & 0.7529 & 0.8548 & 0.2404 & 0.0454 & 0.1017 & 0.7863 & 0.2561 \\
GlyphDraw2\cite{glyphdraw2} & 0.7369 & 0.8921 & - & - & - & - & - & - \\
TextGen\cite{textgen} & 0.7336 & 0.8898 & - & - & 0.6792 & 0.8394 & - & - \\
AnyText\cite{anytext} & 0.7239 & 0.8760 & 0.8841 & - & 0.6923 & 0.8396 & 0.8015 & - \\
AnyText\textcolor{green}{\dag} & 0.7041 & 0.8827 & 0.8827 & 0.2624 & 0.6452 & 0.8181 & 0.8022 & 0.2588 \\
AnyText2\cite{anytext2} & 0.8096 & 0.9184 & 0.8963 & - & 0.7130 & 0.8516 & 0.8139 & - \\
AnyText2\textcolor{green}{\dag} & 0.8122 & 0.9194 & 0.8941 & 0.2497 & 0.7171 & 0.8488 & 0.8102 & 0.2486 \\
\midrule
GRPO baseline & 0.7246 & 0.8935 & 0.8970 & 0.2708 & 0.6782 & 0.8663 & 0.8138 & 0.2668 \\
\rowcolor{gray!20}POCA & 0.7651 & 0.8983 & 0.8985 & 0.2694 & 0.6942 & 0.8696 & 0.8170 & 0.2653 \\
\bottomrule
\end{tabular}%
}
\end{table*}

\section{Experiments}
\begin{figure*}[t]
  \vspace{-1em}
  \centering
  \includegraphics[width=\linewidth]{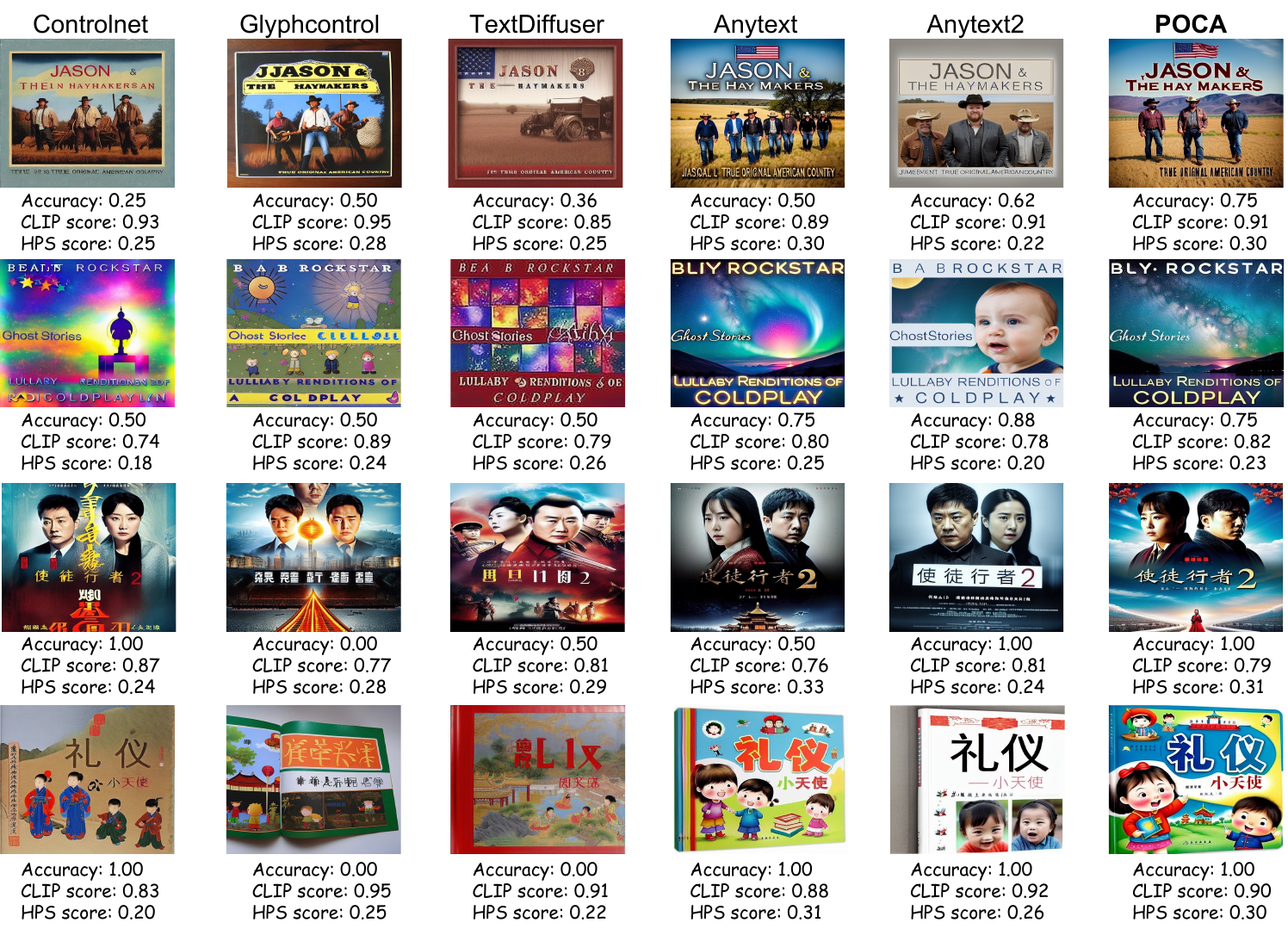}
  \vspace{-2em}
  \caption{\textbf{Qualitative comparison of POCA and other methods on the AnyText-benchmark.} POCA demonstrates overall superior performance in text accuracy, text-image alignment and aesthetic quality.}
  \label{fig:compare_all}
  \vspace{-1em}
\end{figure*}

\begin{figure}[htbp]
  \centering
  \includegraphics[width=\linewidth]{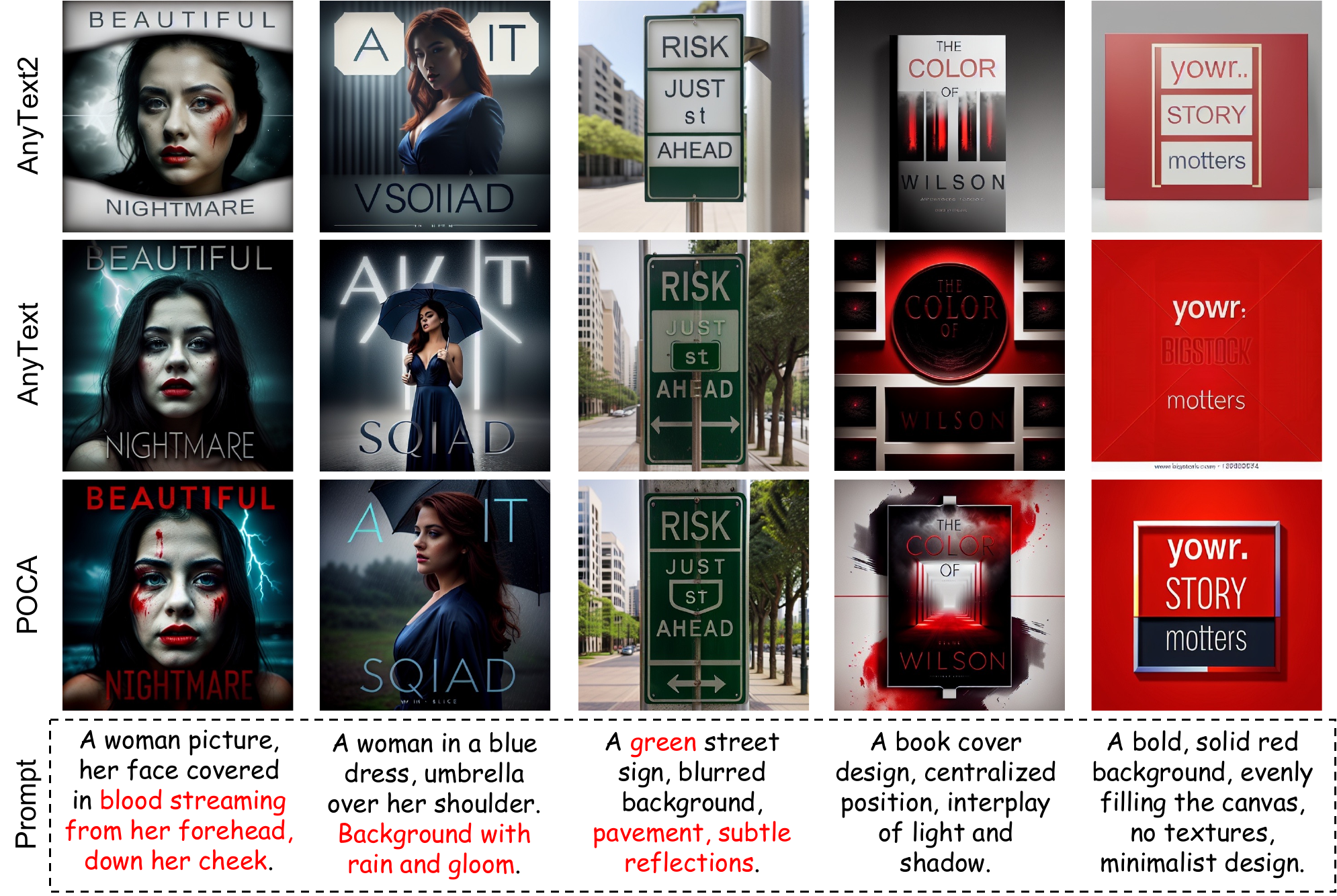}
  \caption{Qualitative comparison of POCA against AnyText and AnyText2 on complex prompts from POCA-20k test set. POCA shows superior capabilities in aesthetic appeal and complex instruction-following.}
  \label{fig:compare_prompt}
  \vspace{-1em}
\end{figure}

\begin{figure*}[ht]
  \centering
  \includegraphics[width=0.9\textwidth]{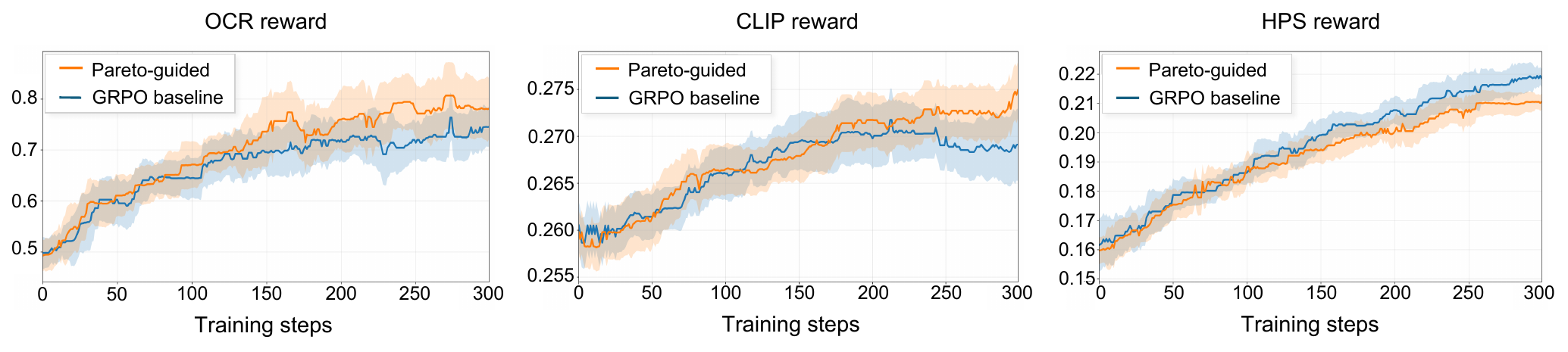}
  \vspace{-0.5em}
  \caption{\textbf{Reward curve comparison.} The GRPO baseline uses weighted sum approach to aggregate OCR, CLIP and HPS scores in a ratio of {1,1,1}, which leads to suboptimal results in the OCR and CLIP rewards compared to our Pareto-guided approach.}
  \label{fig:reward_ablation}
  \vspace{-0.5em}
\end{figure*}

\begin{figure}[ht]
  \centering
  \captionsetup{belowskip=-12pt}
  \includegraphics[width=\linewidth]{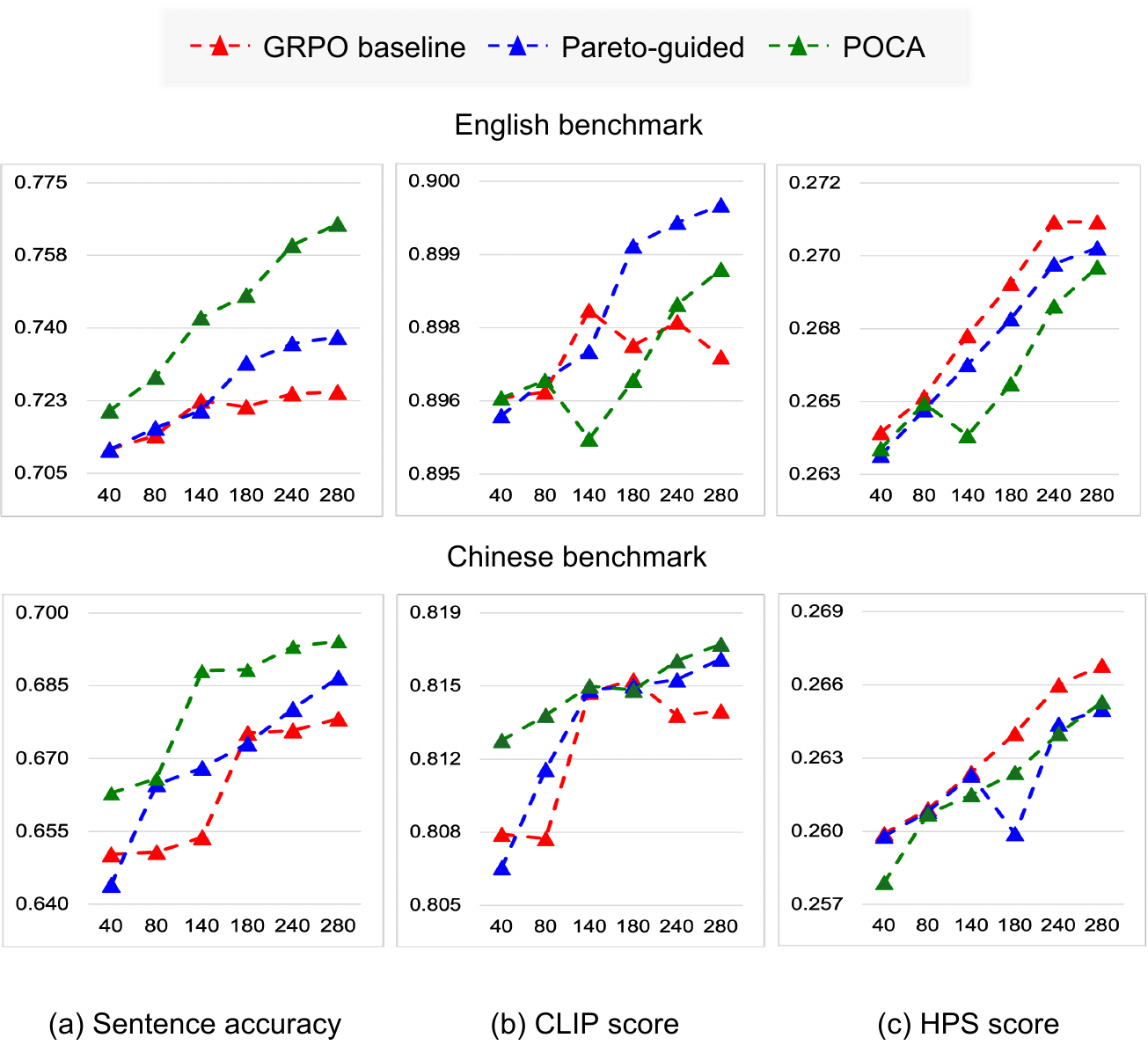}
  \vspace{-1.5em}
  \caption{\textbf{Performance comparisons of POCA, the Pareto-guided baseline, and the GRPO baseline.} The performance of all three methods is evaluated by (a) Sen.ACC, (b) CLIP score and (3) HPS score at training steps from 40 to 280.}
  \label{fig:performance_ablation}
  \vspace{-1em}
\end{figure}

\subsection{Experiment Setting}
\textbf{Dataset:} POCA is designed to align the model's generative ability on text accuracy, aesthetic quality and instruction-following. To provide an informative prompt set, we first mix images from three distinct domains: 1) 5k \textbf{general images} that contain 200 classes of various general objects from SynthText~\cite{synthtext} to maintain a generalized base model. 2) 10k \textbf{scene text images} rich in real-world text are extracted from AnyWord-3M~\cite{anytext} to provide strong signals for optimizing text accuracy. 3) 5k \textbf{art and movie posters} are taken from LeX-Art~\cite{lex_art}, which are biased toward aesthetic and coherence optimization objectives. Subsequently, we use Gemini 2.5~\cite{gemini} to generate descriptive prompts for all 20k images as our training dataset POCA-20k. More details and statistics are in Appendix~\ref{sec:dataset}.

\textbf{Reward Models:} We incorporate three reward models for optimization objectives. 1) The CLIP~\cite{clip} model with the ViT-B/32 image encoder is used to measure text-image alignment. 2) EasyOCR~\cite{easyocr} is implemented as our OCR reward model by first recognizing the generated text and measuring the text accuracy using Normalized Edit Distance (NED) with the target text, as shown in Appendix~\ref{sec:ned}. 3) HPS-v2.1~\cite{hpsscore} is applied as our aesthetic scorer to quantify visual appeal of generated text image. It is pretrained on human-rated data.

\textbf{Implementation Details:} We adopt AnyText~\cite{anytext} as our base visual text generation model since it is an open-source and widely used baseline. We strictly follow DanceGRPO~\cite{dancegrpo} training configuration with a batch size of 4 and 50 denoising steps to sample each prompt 16 times without classifier-free guidance. Instead of updating all parameters, we use LoRA with a rank of 128. Finally, the base model is trained for 300 steps using POCA-20k.

\subsection{Quantitative Evaluation}
\label{sec:quan_eval}
We conduct our main evaluation on the AnyText-benchmark~\cite{anytext}, which is widely adopted by visual text generation models. To ensure a fair comparison, we strictly follow the evaluation settings proposed by AnyText~\cite{anytext}. We evaluate all methods on three key criteria: 1) \textbf{Text Accuracy} using sentence accuracy (Sen.ACC) and normalized edit distance (NED), 2) \textbf{Text-Image Alignment} with CLIP score, and 3) \textbf{Aesthetic Quality} with HPS score. We first compare against leading visual text generation models~\cite{controlnet,textdiffuser,glyphcontrol,glyphdraw2,textgen,anytext,anytext2}. For publicly available methods, we use their official settings. For others, we reference the metrics reported in their papers. Moreover, we also compare against a GRPO baseline, which applies the standard GRPO objective (Eq.~\ref{eq:dancegrpoloss}) to our dataset using an equally-weighted scalarization of all rewards. As shown in Table~\ref{table:performance_comparison}, POCA substantially improves upon the original AnyText base model across all metrics and surpasses the performance of the current SOTA model, AnyText 2~\cite{anytext2}, in terms of text-image alignment and aesthetics. Notably, POCA also outperforms the GRPO baseline in both text accuracy and CLIP score, which demonstrates that our proposed framework successfully finds a superior balance between conflicting rewards, overcoming the limitations of scalarization. We also validate POCA on a more recent SDXL-based model~\cite{glyphxl} and observe similar improvements, indicating that our framework generalizes well to larger models. The detailed results are provided in the Appendix~\ref{sec:largermodel}.

\vspace{-1em}
\begin{figure}[H]
    \centering
    \includegraphics[width=0.8\linewidth]{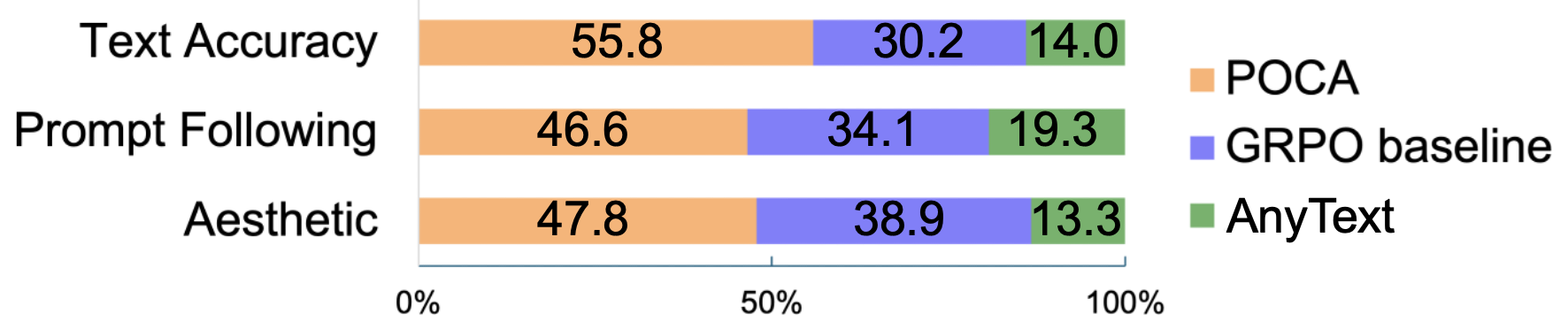}
    \vspace{-0.6em}
    \caption{Human study results in terms of text accuracy, prompt following, and aesthetics, measured by human preference win rates.}
    \vspace{-0.5em}
    \label{fig:human_study}
\end{figure}

\subsection{Qualitative Analysis}
\textbf{Comparison with Baselines:} We further demonstrate the effectiveness of POCA on both Chinese and English prompts, using test cases from the AnyText-benchmark~\cite{anytext}. Fig.~\ref{fig:compare_all} provides a visual comparison between POCA and multiple baselines, alongside their corresponding reward scores. POCA achieves higher overall reward scores and produces visually superior images, particularly in aspects like seamless text blending, harmonious color combinations, and finer image details. This improvement can be attributed to our base model being aligned with multiple rewards simultaneously with filtered unambiguous reward signals and an efficient learning path.

\textbf{Aesthetic Appeal and Instruction-Following:} We then conduct a more challenging qualitative analysis to specifically evaluate POCA's ability to handle complex prompt instructions. We curated a new test set by randomly sampling 50 untouched prompts from our POCA-20k dataset. As noted, these prompts contain more detailed instructions. Fig.~\ref{fig:compare_prompt} provides a visual comparison with AnyText~\cite{anytext} and AnyText2~\cite{anytext2} on five of these complex prompts. It can be observed that the baseline models show weaker instruction-following capabilities compared to POCA. This is particularly evident in their failure to follow fine-grained details. Furthermore, when generating design-like images, they often fail to properly integrate the text with the background, whereas POCA produces a more coherent and aesthetically pleasing result with accurate text.

\textbf{Human Study:} The user study involved 22 participants, who evaluated 50 image sets generated by POCA, AnyText, and GRPO along three dimensions: text accuracy, prompt following, and aesthetics. As shown in Fig.~\ref{fig:human_study}, POCA significantly outperforms the base model and the GRPO baseline across all categories.

\subsection{Ablation Study}
\textbf{Weighted-sum vs. Pareto-guided:} We begin our ablation by validating the central hypothesis of our work that the naive weighted sum of conflicting rewards is the primary cause of instability and suboptimal convergence. The GRPO baseline and our Pareto-guided approach are compared directly without the adaptive curriculum in the same data environment. Their growth (reward) curves in Fig.~\ref{fig:reward_ablation} show that the GRPO baseline converges to a suboptimal state, performing competitively only on the HPS reward, while our Pareto-guided method successfully finds a superior trade-off, achieving significantly higher OCR and CLIP rewards. We further analyze their performance at fixed training snapshots to visualize the convergence path. As shown in Fig.~\ref{fig:performance_ablation}, the GRPO baseline exhibits unstable improvements during multi-reward optimization, particularly in text-image alignment. In contrast, our Pareto-guided approach demonstrates stable and consistent gains in both text accuracy and text-to-image alignment, indicating that bi-directional selection provides a cleaner optimization path by involving only Pareto-guided positive and negative trade-off signals. We also discuss the more recent weighted-sum approach RPO~\cite{weighted-sum} and other Pareto sets in Appendix~\ref{sec:more_comparisons} and ~\ref{sec:pareto_selection}, respectively.
\vspace{-0.5em}
\begin{table}[htbp]
\caption{\small Ablation study of curriculum planning strategy on AnyText-benchmark (English). We analyze the impact of 1) data scalability on subsets of 1k, 5k, and 10k prompts and 2) our adaptive multi-stage design (compared to the one-stage baseline).}
\vspace{-6pt}
\label{table:ablation_curriculum}
\centering
\small
\resizebox{0.85\linewidth}{!}{%
\begin{tabular}{@{}lcccc@{}}
\toprule
\multirow{2}{*}{Setting} & \multicolumn{4}{c}{English} \\
\cmidrule(l){2-5}
 & Sen.ACC$\uparrow$ & NED$\uparrow$ & CLIP score$\uparrow$ & HPS score$\uparrow$ \\
\midrule
Pareto-guided     & 0.7378 & 0.8923 & 0.8996 & 0.2700 \\
POCA-1k           & 0.7374 & 0.8941 & 0.8966 & 0.2678 \\
POCA-5k           & 0.7530 & 0.8968 & 0.8980 & 0.2682 \\
POCA-10k          & 0.7651 & 0.8983 & 0.8985 & 0.2694 \\
POCA-one stage    & 0.7572 & 0.8980 & 0.8973 & 0.2688 \\
\bottomrule
\end{tabular}%
}
\vspace{-0.5em}
\end{table}

\textbf{Adaptive Curriculum vs. Naive Mixing:} As shown in Fig.~\ref{fig:performance_ablation}, the snapshots indicate that our OCR-oriented curriculum yields faster convergence and higher sentence accuracy without sacrificing other reward objectives. This result supports the claim in Sec.~\ref{sec:intro} that adaptive curriculum planning is an effective strategy for data selection. Table~\ref{table:ablation_curriculum} further verifies two key properties of our design: scalability and the benefit of adaptive staging. Specifically, we compare models trained for 300 steps on randomly sampled subsets of 1k, 5k, and 10k images from the English benchmark. POCA-1k achieves performance comparable to that of the Pareto-guided method trained on 10k images, indicating strong data efficiency. Moreover, the multi-stage design (row 4) achieves higher sentence accuracy than the single-stage variant (row 5), further justifying this design.

%% file: sec/6_conclusion.tex
\vspace{-5pt}
\section{Conclusion}

In this work, we addressed the suboptimal convergence of GRPO-based methods caused by naive reward scalarization and the practical challenge of inefficient data selection by introducing POCA. By removing ambiguous optimization signals using our Pareto set sorting, POCA stabilizes multi-reward optimization and generates higher-quality text images with improved aesthetics and instruction-following ability. Furthermore, our curriculum planning strategy enables faster convergence and better performance through a self-paced learning path.

\par\noindent\textbf{Acknowledgments} This work was supported in part by JST, CRONOS, Japan Grant Number JPMJCS24K4.

%% file: sec/X_suppl.tex
\clearpage
\setcounter{page}{1}
\maketitlesupplementary

%\section{Rationale}
%\label{sec:rationale}
%\begin{itemize}
%\item The supplementary can back-reference sections of the main paper, for example, we can refer to \cref{sec:intro};
%\item The main paper can forward reference sub-sections within the supplementary explicitly (e.g. referring to a particular experiment); 
%\item When submitted to arXiv, the supplementary will already included at the end of the paper.
%\end{itemize}
% 
%To split the supplementary pages from the main paper, you can use \href{https://support.apple.com/en-ca/guide/preview/prvw11793/mac#:~:text=Delete%20a%20page%20from%20a,or%20choose%20Edit%20%3E%20Delete).}{Preview (on macOS)}, \href{https://www.adobe.com/acrobat/how-to/delete-pages-from-pdf.html#:~:text=Choose%20%E2%80%9CTools%E2%80%9D%20%3E%20%E2%80%9COrganize,or%20pages%20from%20the%20file.}{Adobe Acrobat} (on all OSs), as well as \href{https://superuser.com/questions/517986/is-it-possible-to-delete-some-pages-of-a-pdf-document}{command line tools}.

\section{Normalized Edit Distance}
\label{sec:ned}
Edit Distance (ED), also known as Levenshtein distance, measures the minimum number of operations required to transform one string into another. The allowed operations include character insertions, deletions, and substitutions, each contributing a unit cost to the total edit distance. This metric is widely used for text similarity evaluation. Normalized Edit Distance (NED) normalizes the computed ED by dividing it by the maximum length of the two strings, ensuring a value between 0 and 1, where 0 indicates identical strings and 1 represents completely different strings. The formal computation of NED is shown in Algorithm \ref{alg:ned}, where a two-dimensional dynamic programming (DP) table is used to iteratively compute the minimum edit cost between two input strings. In this work, we apply NED as our OCR reward, since Sen.ACC suffers from reward sparsity, which typically assigns a zero score to partially correct generations, failing to provide the fine-grained feedback necessary for optimization. In contrast, NED offers a continuous measure of character-level alignment. To align this distance metric with the standard RL objective of reward maximization, we formulate the training reward as $1 - \text{NED}$.

\begin{algorithm}
\caption{Normalized Edit Distance (NED)}
\label{alg:ned}
\footnotesize
\begin{algorithmic}[1]
\State Given two strings $a$ and $b$ of lengths $n$ and $m$. Initialize DP table $D \in \mathbb{R}^{(n+1) \times (m+1)}$. Define edit cost $c$. Initialize $D_{i,0} \gets i$, $D_{0,j} \gets j$ for all $i, j$.
\For{$i = 1$ to $n$}
    \For{$j = 1$ to $m$}
        \State $D_{i,j} \gets \min(D_{i-1,j} + 1, D_{i,j-1} + 1, D_{i-1,j-1} + I(a_i \neq b_j))$
    \EndFor
\EndFor
\State \Return $D_{n,m} / \max(n, m)$
\end{algorithmic}
\end{algorithm}

\vspace{-1em}
\section{Pareto Set Comparison}
\label{sec:pareto_selection}
\begin{figure*}[ht]
  \centering
  \captionsetup{aboveskip=-1pt, belowskip=0pt}
  \includegraphics[width=0.9\textwidth]{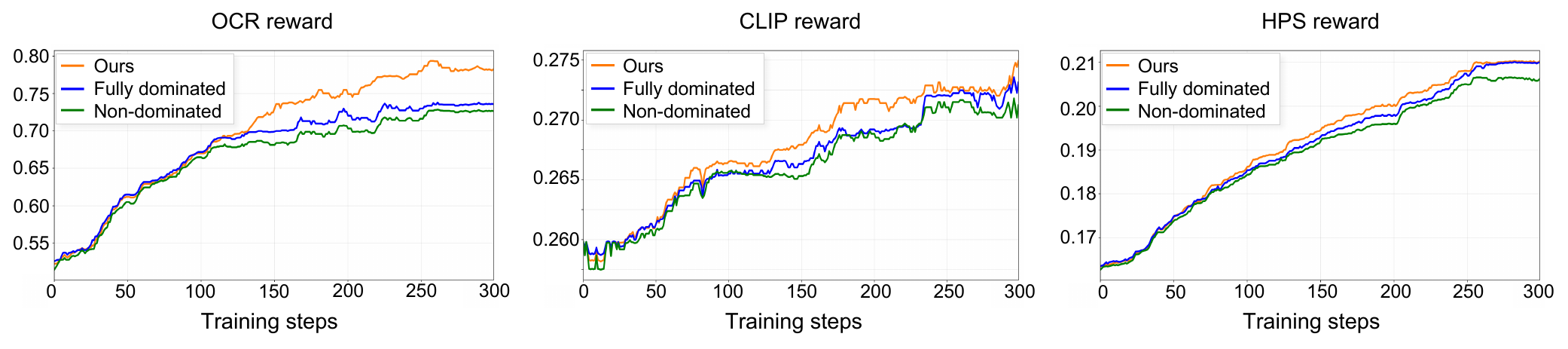}
  \caption{\textbf{Reward curves for different Pareto selection algorithms.} The figure illustrates that our bi-directional approach consistently achieves higher reward curves across all three metrics compared to one-direction baselines. This confirms the superior effectiveness of using both positive and negative signals for stable and efficient policy optimization.}
  \label{fig:supp_reward_ablation}
\end{figure*}

To validate the effectiveness of our bi-directional strategy, we compare it against Parrot~\cite{parrot}, which employs one-directional non-dominated sorting to update the policy using only the best samples. Additionally, we investigate the impact of negative samples by evaluating a fully dominated sorting baseline. Following the configuration in~\cite{dancegrpo}, we train the base model using each sorting strategy on the POCA-20k dataset for 300 steps. Fig.~\ref{fig:supp_reward_ablation} illustrates that our bi-directional sorting algorithm consistently achieves higher reward curves across all three metrics compared to one-direction baselines. Table~\ref{table:pareto_comparison} further validates the superiority of our approach, demonstrating the overall best performance in Sen.ACC, CLIP score, and HPS score. The combined evidence from the training curves and quantitative metrics confirms that leveraging both positive and negative signals is essential for achieving the best convergence and overall performance in multi-reward policy optimization.

\begin{figure}[ht]
  \centering
  \captionsetup{belowskip=-5pt}
  \includegraphics[width=\linewidth]{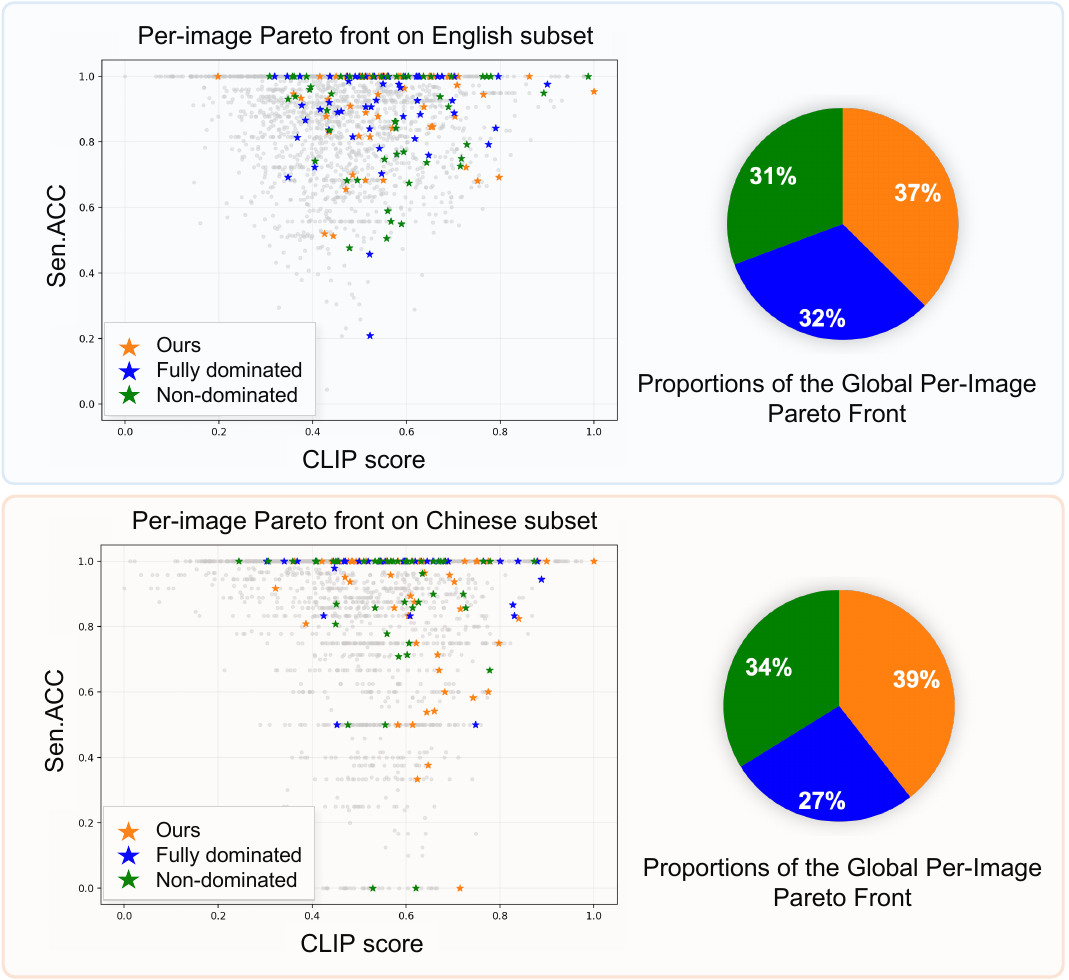}
  \caption{\small \textbf{Contribution to the global Pareto front.} The figure shows the fraction of globally non-dominated solutions contributed by each method in the unified solution space. Our bi-directional approach accounts for the largest share of optimal solutions, indicating a stronger coverage of the Pareto front than one-direction baselines.}
  \label{fig:supp_pareto_front}
  \vspace{-1em}
\end{figure}

To further assess the upper-bound capability of the learned Pareto sets, we evaluate them using a best@$k$ protocol. In detail, we randomly sample 100 prompts from both English and Chinese benchmarks and generate $k=8$ candidates per prompt using each method. We then apply a rule-based selection strategy that prioritizes Sen.ACC, followed by CLIP score and finally HPS score, to pick the best image among all candidates. Table~\ref{table:pareto_comparison_best@8} indicates that our method has a higher probability of producing the optimal trade-off state in a larger solution space. Finally, we compare all methods in a unified global solution space of their generated solutions. We identify the global Pareto front by calculating the non-dominated points per-image across this combined set and quantify the relative contribution of each method, as shown in Fig.~\ref{fig:supp_pareto_front}. For both English and Chinese subsets, our bi-directional approach dominates the global Pareto front, accounting for the largest proportion of non-dominated solutions.

\begin{table*}[t]
\centering
\footnotesize
\caption{\small Quantitative comparison of different Pareto selection strategies.}
\label{tab:pareto_all}

% ---------- (a) Full validation set ----------
\begin{subtable}{0.9\textwidth}
  \centering
  \caption{\small \textbf{Averaged performance of Pareto sorting strategies.} The table compares the averaged performance of one-direction methods against our bi-directional approach, validating that using both positive and negative signals leads to superior overall alignment across all metrics.}
  \label{table:pareto_comparison}
  \resizebox{\textwidth}{!}{%
    \begin{tabular}{@{}lcccc|cccc@{}}
      \toprule
      \multirow{2}{*}{Methods} &
      \multicolumn{4}{c|}{English} &
      \multicolumn{4}{c}{Chinese} \\
      \cmidrule(l){2-5} \cmidrule(l){6-9}
       & Sen.ACC$\uparrow$ & NED$\uparrow$ & CLIP score$\uparrow$ & HPS score$\uparrow$
       & Sen.ACC$\uparrow$ & NED$\uparrow$ & CLIP score$\uparrow$ & HPS score$\uparrow$ \\
      \midrule
      GRPO baseline      & 0.7246 & 0.8935 & 0.8970 & 0.2708 & 0.6782 & 0.8663 & 0.8138 & 0.2668 \\
      Non-dominated set  & 0.7227 & 0.8827 & 0.8990 & 0.2710 & 0.6758 & 0.8655 & 0.8161 & 0.2674 \\
      Fully dominated set& 0.7274 & 0.8836 & 0.9019 & 0.2712 & 0.6766 & 0.8656 & 0.8154 & 0.2671 \\
      \midrule
      \textbf{Ours}      & \textbf{0.7378} & \textbf{0.8923} & \textbf{0.8996} & \textbf{0.2700} &
                           \textbf{0.6867} & \textbf{0.8666} & \textbf{0.8163} & \textbf{0.2650} \\
      \bottomrule
    \end{tabular}%
  }
\end{subtable}

\vspace{6pt}

% ---------- (b) best@8 on 100-sample subset ----------
\begin{subtable}{0.9\textwidth}
  \centering
  \caption{\small \textbf{Evaluation of upper-bound generation capability.} This table demonstrates that our bi-directional selection effectively expands the Pareto frontier, leading to a higher probability of discovering optimal trade-off solutions.}
  \label{table:pareto_comparison_best@8}
  \resizebox{\textwidth}{!}{%
    \begin{tabular}{@{}lcccc|cccc@{}}
      \toprule
      \multirow{2}{*}{Methods} &
      \multicolumn{4}{c|}{English} &
      \multicolumn{4}{c}{Chinese} \\
      \cmidrule(l){2-5} \cmidrule(l){6-9}
       & Sen.ACC$\uparrow$ & NED$\uparrow$ & CLIP score$\uparrow$ & HPS score$\uparrow$
       & Sen.ACC$\uparrow$ & NED$\uparrow$ & CLIP score$\uparrow$ & HPS score$\uparrow$ \\
      \midrule
      Non-dominated set   & 0.8418 & 0.9220 & 0.9439 & 0.2760 & 0.8769 & 0.9496 & 0.8580 & 0.2702 \\
      Fully dominated set & 0.8418 & 0.9223 & 0.9467 & 0.2762 & 0.8769 & 0.9475 & 0.8623 & 0.2725 \\
      \midrule
      \textbf{Ours}       & \textbf{0.8491} & \textbf{0.9204} & \textbf{0.9438} & \textbf{0.2766} &
                            \textbf{0.8821} & \textbf{0.9495} & \textbf{0.8640} & \textbf{0.2741} \\
      \bottomrule
    \end{tabular}%
  }
\end{subtable}

\end{table*}

\section{Variance Analysis of Reward Models}
\label{sec:variance_compare}
\begin{figure}[ht]
  \centering
  \captionsetup{belowskip=-10pt}
  \includegraphics[width=\linewidth]{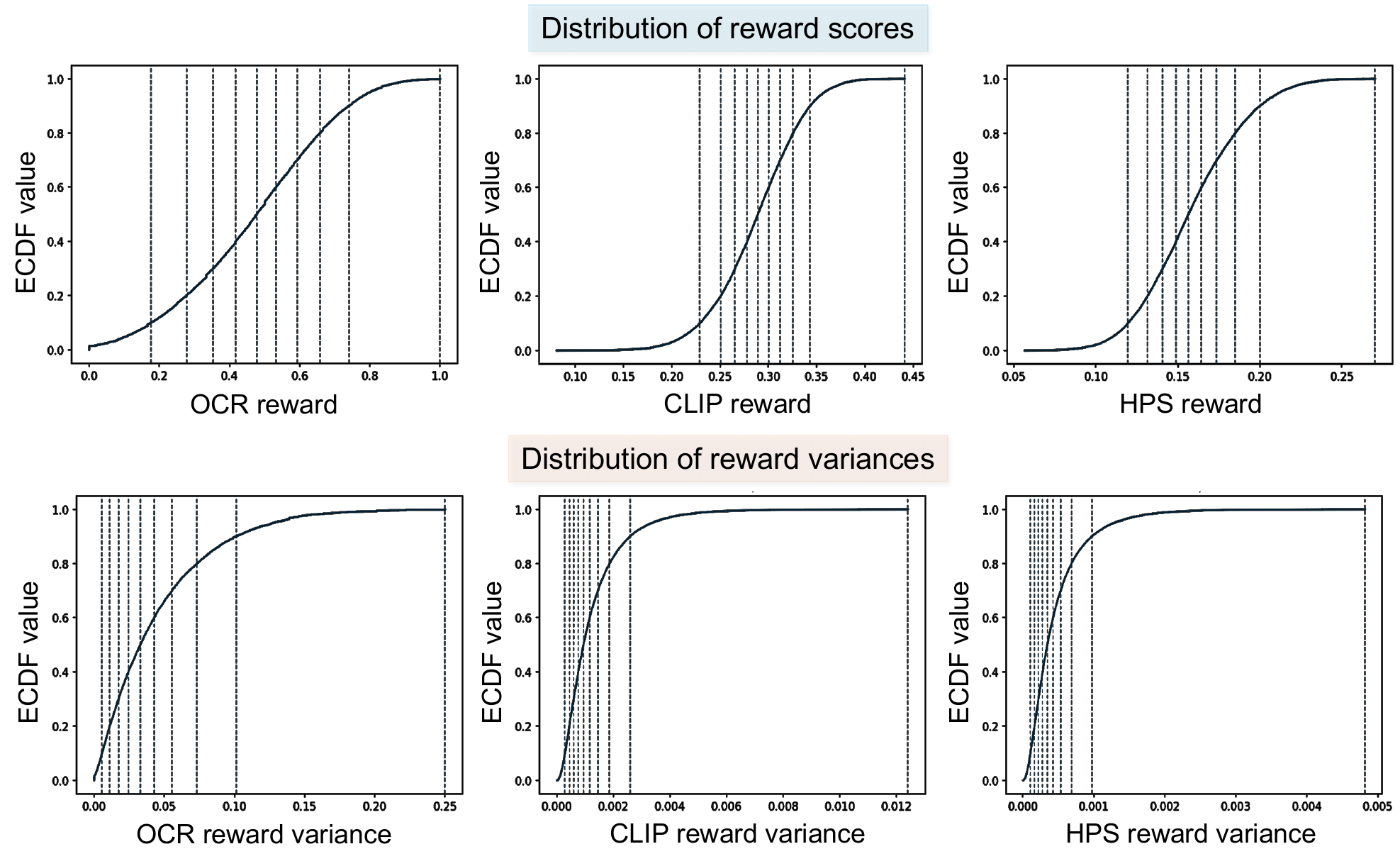}
  \caption{\small ECDFs for both the mean reward scores and the reward variances of OCR, CLIP, and HPS across the full training set. Vertical dashed lines indicate deciles (10\%–100\% in steps of 10\%).}
  \label{fig:variance_compare}
  \vspace{-1em}
\end{figure}

To justify our choice of the OCR reward as the difficulty measure in the curriculum, we compare the distributions of all three reward signals (OCR, CLIP, and HPS) over the full training set. For each reward model, we compute scores for every sample and plot the ECDFs together with selected quantiles, as shown in Fig.~\ref{fig:variance_compare}. We observe that CLIP and HPS scores are highly concentrated in a narrow band between the 20th and 80th percentiles, leading to a very limited dynamic range in the central region of the dataset. In contrast, the OCR reward spans a much broader interval over the same percentile range, and its inter-percentile gaps (e.g., at the 20th, 40th, 60th, and 80th percentiles) are consistently larger than those of CLIP and HPS, which is also reflected by a noticeably larger variance. This indicates that OCR provides greater variance and finer ranking resolution across samples. These findings support our design choice that OCR reward provides a more informative and discriminative difficulty signal for curriculum scheduling.

\begin{table*}[t]
    \centering
    % =========== Performance Comparison ===========
    \caption{\small Comparisons between POCA and the counterparts.}
    \vspace{-10pt}
    \label{table:rebuttal_performance_comparison}
    \footnotesize
    \resizebox{0.9\linewidth}{!}{% 
    \begin{tabular}{@{}lcccc|cccc@{}}
    \toprule
    \multirow{2}{*}{Methods} & 
    \multicolumn{4}{c|}{English} & 
    \multicolumn{4}{c}{Chinese} \\
    \cmidrule(l){2-5} \cmidrule(l){6-9}
     & Sen.ACC$\uparrow$ & NED$\uparrow$ & CLIP score$\uparrow$ & HPS score$\uparrow$ & Sen.ACC$\uparrow$ & NED$\uparrow$ & CLIP score$\uparrow$ & HPS score$\uparrow$ \\
    \midrule
    RPO-Harmonic & 0.7400 & 0.8875 & \textbf{0.9029} & 0.2678 & 0.6908 & 0.8684 & 0.8155 & \textbf{0.2672} \\
    Curriculum-DPO & 0.7268 & 0.8866 & 0.8962 & 0.2602 & 0.6574 & 0.8435 & 0.8011 & 0.2598 \\
    \textbf{POCA} & \textbf{0.7651} & \textbf{0.8983} & 0.8985 & \textbf{0.2694} & \textbf{0.6942} & \textbf{0.8696} & \textbf{0.8170} & 0.2653 \\
    \midrule
    Glyph-SDXL-v2 & 0.5950 & 0.7452 & 0.8553 & 0.2131 & 0.6174 & 0.7608 & 0.7796 & 0.2131 \\
    Pareto-guided-SDXL & 0.6119 & 0.7635 & 0.8700 & 0.2293 & 0.6653 & 0.8281 & \textbf{0.8042} & 0.2294 \\
    \textbf{POCA-SDXL} & \textbf{0.6218} & \textbf{0.7775} & \textbf{0.8762} & \textbf{0.2302} & \textbf{0.6815} & \textbf{0.8371} & 0.7996 & \textbf{0.2332} \\
    \bottomrule
    \end{tabular}%
    }
\end{table*}

\section{More Details About Dataset Preparation}
\label{sec:dataset}
\begin{figure}[htpb]
  \centering
  \captionsetup{aboveskip=3pt, belowskip=5pt}
  \includegraphics[width=\linewidth]{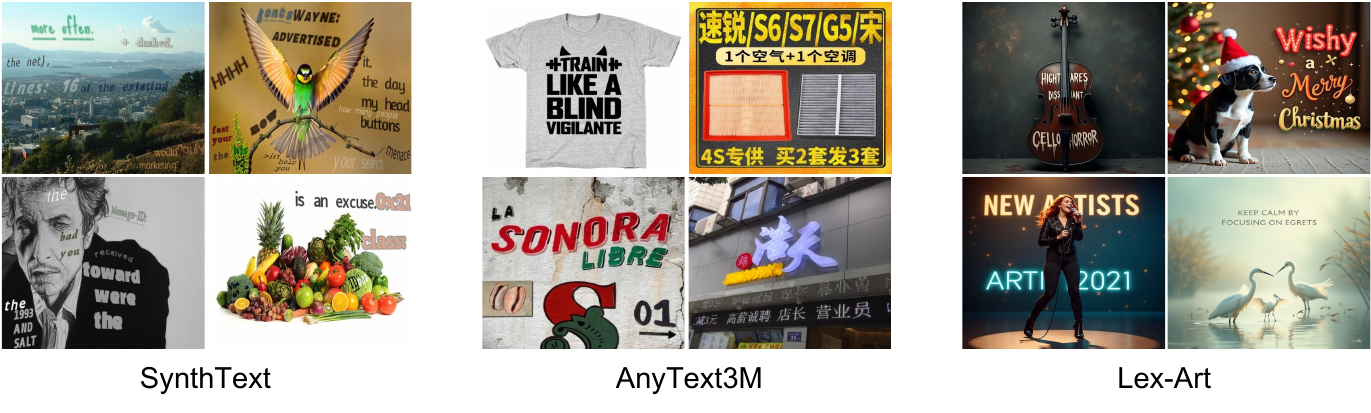}
  \caption{Examples of the diverse image domains in our image dataset.}
  \vspace{-2em}
  \label{fig:supp_dataset}
\end{figure}

\begin{figure}[t]
  \centering
  \captionsetup{aboveskip=3pt, belowskip=5pt}
  \includegraphics[width=\linewidth]{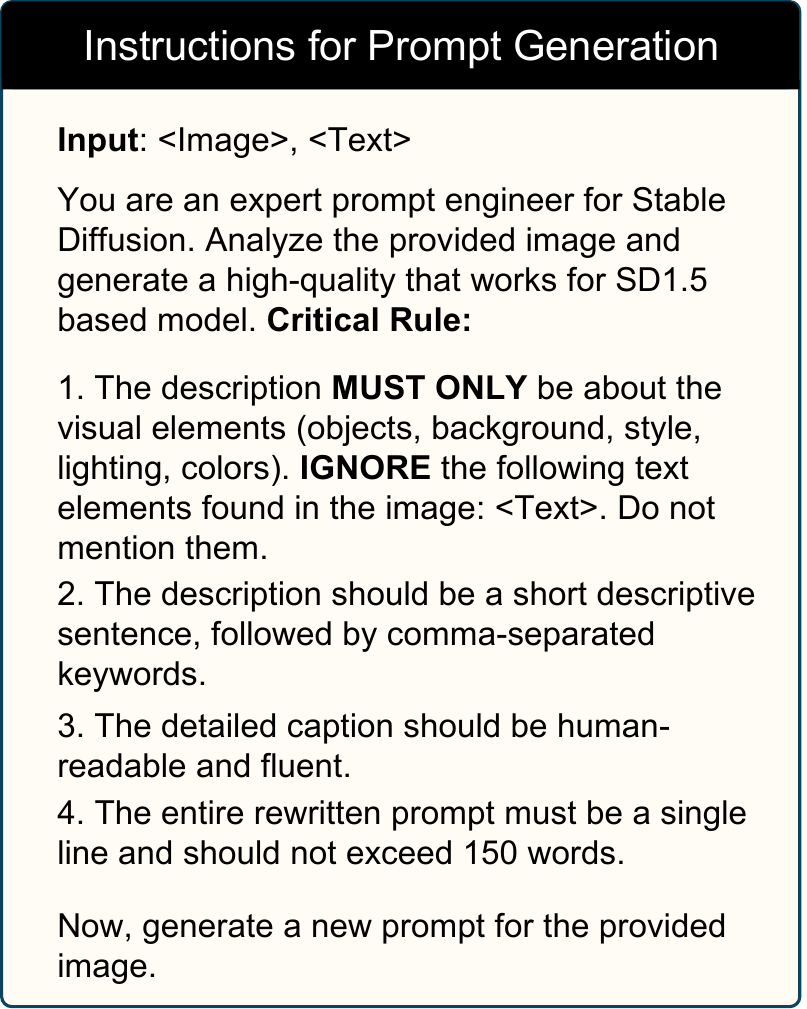}
  \caption{Rule-based instructions utilized with the Gemini 2.5 model to generate high-fidelity prompts for the POCA-20k dataset.}
  \label{fig:supp_prompt_generation}
\end{figure}

Our image dataset is randomly sampled from the following datasets:
\begin{itemize}
    \item SynthText~\cite{synthtext} renders synthetic English text onto real-world background images spanning 200 classes, including human portraits, animals, and natural landscapes. Text is composited into these backgrounds using a rule-based rendering engine, producing diverse text images. 
    
    \item AnyWord-3M~\cite{anytext} is a large-scale multilingual dataset extracted from publicly available images. These images cover a wide range of real-scene images containing text, including street views, advertisements and book covers.
    
    \item LeX-10K~\cite{lex_art} is a curated collection of 10K English text–image pairs tailored for visual text generation, with a strong emphasis on aesthetics, text fidelity, and stylistic diversity. The images cover a wide range of layouts and themes, such as posters, logos and design-like images.
\end{itemize}
We sampled 5k images from SynthText~\cite{synthtext}, which contains all 200 classes, 10k images from AnyText-3M~\cite{anytext} with 5k English text images and 5k Chinese text images, and 5k images from LeX-10K~\cite{lex_art}. Examples of collected images are shown in Fig.~\ref{fig:supp_dataset}. For training prompts, we leveraged Gemini 2.5~\cite{gemini} to describe each image and Fig.~\ref{fig:supp_prompt_generation} illustrates the instruction used for prompt generation.

\section{POCA on Larger Model}
\label{sec:largermodel}
Focusing on visual text generation, we use the state-of-the-art AnyText for our main experiments. To demonstrate that POCA is model-agnostic, we also evaluate it using the more recent Glyph-SDXL-v2. Table~\ref{table:rebuttal_performance_comparison} shows the results. Obviously, using Bi-directional Pareto sorting (Pareto-guided-SDXL) can significantly improve the performance on all metrics and introducing curriculum learning (POCA-SDXL) results in an additional improvement. Note that the base models of Glyph-SDXL-v2 and AnyText are trained using different datasets, hence their performances vary. Nevertheless, applying POCA to them leads to remarkable improvements, which indicates that POCA is effective and general.

\vspace{-0.5em}

\section{More Comparisons with Related Works}
\label{sec:more_comparisons}
In this section, we further compare POCA with additional related methods, including 1) the weighted-sum approach RPO~\cite{weighted-sum} and 2) the DPO-based curriculum design, Curriculum-DPO~\cite{curriculumDPO}. 

RPO proposes using the harmonic mean instead of a naive weighted-sum approach to aggregate different rewards in a two-reward setting, so as to place greater emphasis on smaller rewards. In other words, a high final reward is obtained only when both rewards are relatively large. We extend the harmonic reward function in RPO to support three rewards: $r = \frac{1}{\frac{\lambda}{r_{ocr}} + \frac{\alpha}{r_{clip}} + \frac{\beta}{r_{hps}}}$, where $\lambda = \alpha = \beta = \frac{1}{3}$. As shown in Table~\ref{table:rebuttal_performance_comparison}, POCA outperforms RPO on multiple metrics, especially Sen.ACC. Unlike RPO and other weighted-sum methods, POCA avoids the difficulty of balancing aggregation hyperparameters. 

While Curriculum-DPO builds an easy-to-hard learning path by ranking candidate samples with a single reward model and progressively training on preference pairs from coarse, easily distinguishable pairs to finer, harder ones, POCA is designed to address multiple conflicting rewards, leading to distinct sample ranking strategies. We compared POCA with Curriculum-DPO by using OCR rewards for pair ranking. As shown in Table~\ref{table:rebuttal_performance_comparison}, Curriculum-DPO surpasses the AnyText baseline and performs similarly to GPRO, but remains inferior to POCA, particularly across multiple metrics.

\vspace{-0.5em}

\section{Assessment of computational overhead:} Generating the 20k training prompts using Gemini 2.5 took $\sim$40 hours. Performing inference on the entire set of prompts for difficulty measurement requires $\sim$15 hours with 8 GPUs. 

\section{More Visual Examples}
\begin{figure*}[ht]
  \centering
  \captionsetup{aboveskip=3pt, belowskip=5pt}
  \includegraphics[width=\textwidth]{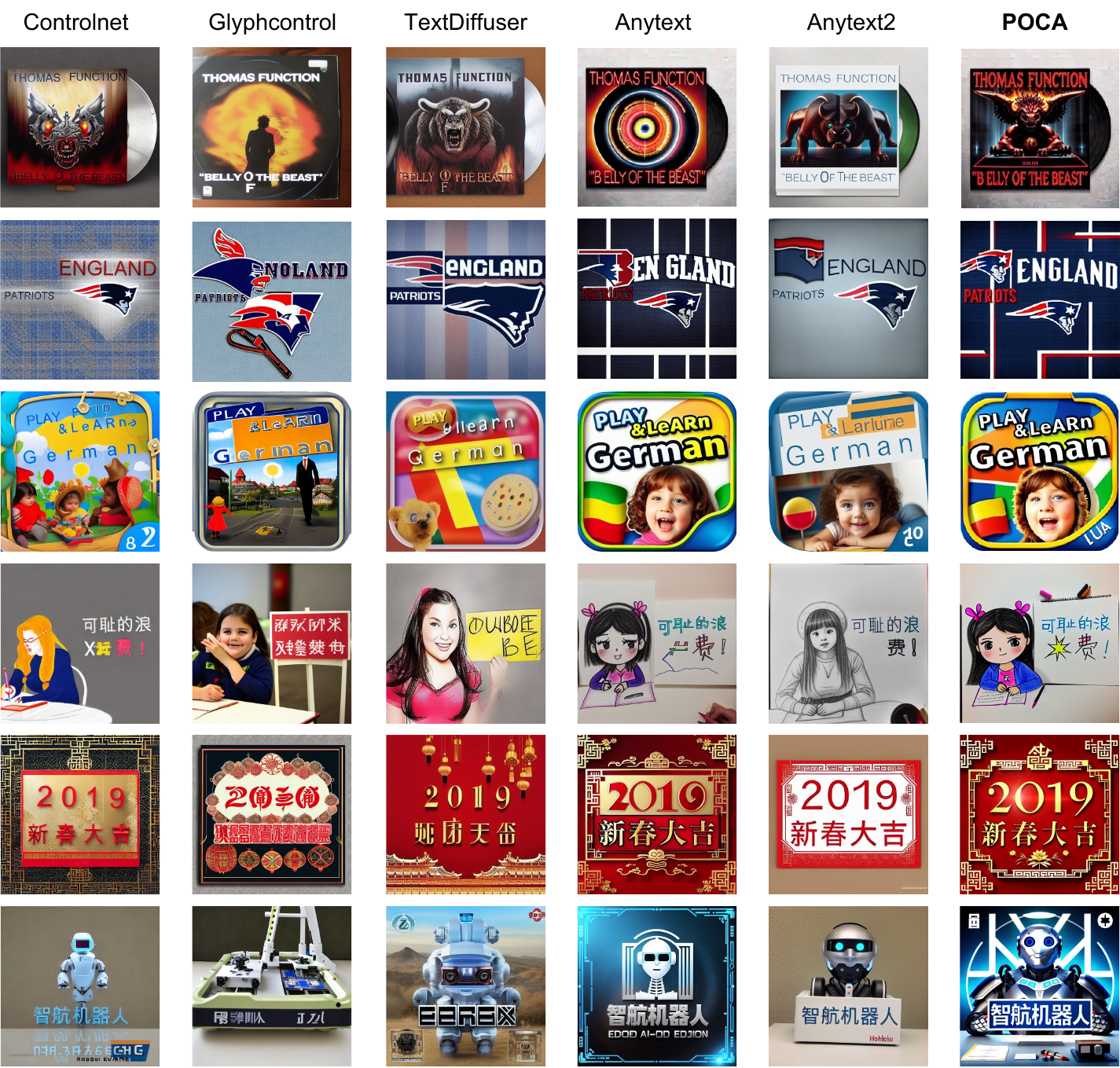}
  \caption{\textbf{General qualitative comparison of POCA and other baselines.} The figure demonstrates POCA's superior overall balance between text accuracy, image coherence, and aesthetic appeal compared to state-of-the-art methods.}
  \label{fig:supp_more_quality}
\end{figure*}

\begin{figure*}[ht]
  \centering
  \captionsetup{aboveskip=3pt, belowskip=5pt}
  \includegraphics[width=\textwidth]{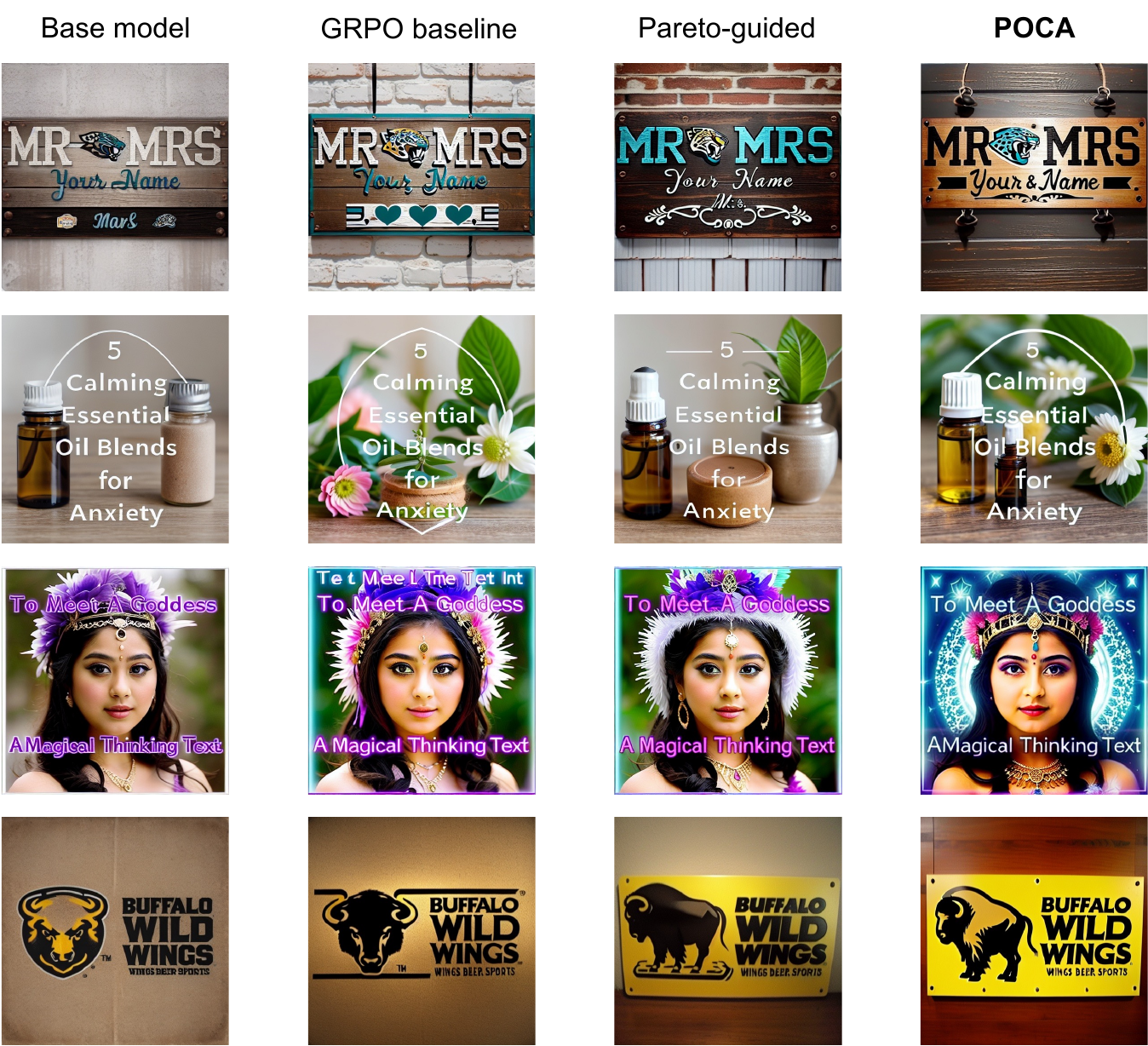}
  \caption{\textbf{Qualitative comparison with standard GRPO baseline.} Although the GRPO baseline improves both aesthetics and text accuracy to some extent, the inconsistent reward signals lead to visually unbalanced images with excessive text and semantic inconsistencies.}
  \label{fig:supp_grpo_compare}
\end{figure*}

We show additional visual examples of POCA in this section. We first provide more examples for comparison with other baselines in Fig.~\ref{fig:supp_more_quality}. These images illustrate that POCA maintains image coherence while generating accurate text, allowing the text to seamlessly integrate with the background. Fig.~\ref{fig:supp_grpo_compare} shows the comparison with the standard weighted-sum GRPO method. While the GRPO baseline brings some degree of improvement in accuracy and aesthetics to the base model, it struggles to properly balance multiple reward objectives. For example, it often generates semantically inconsistent superfluous elements and extra text. Conversely, our Pareto-guided method improves image aesthetics by increasing the level of visual detail while remaining faithful to the semantics. POCA further improves the accuracy and clarity of the text.

\begin{figure*}[ht]
  \centering
  \captionsetup{aboveskip=3pt, belowskip=5pt}
  \includegraphics[width=\textwidth]{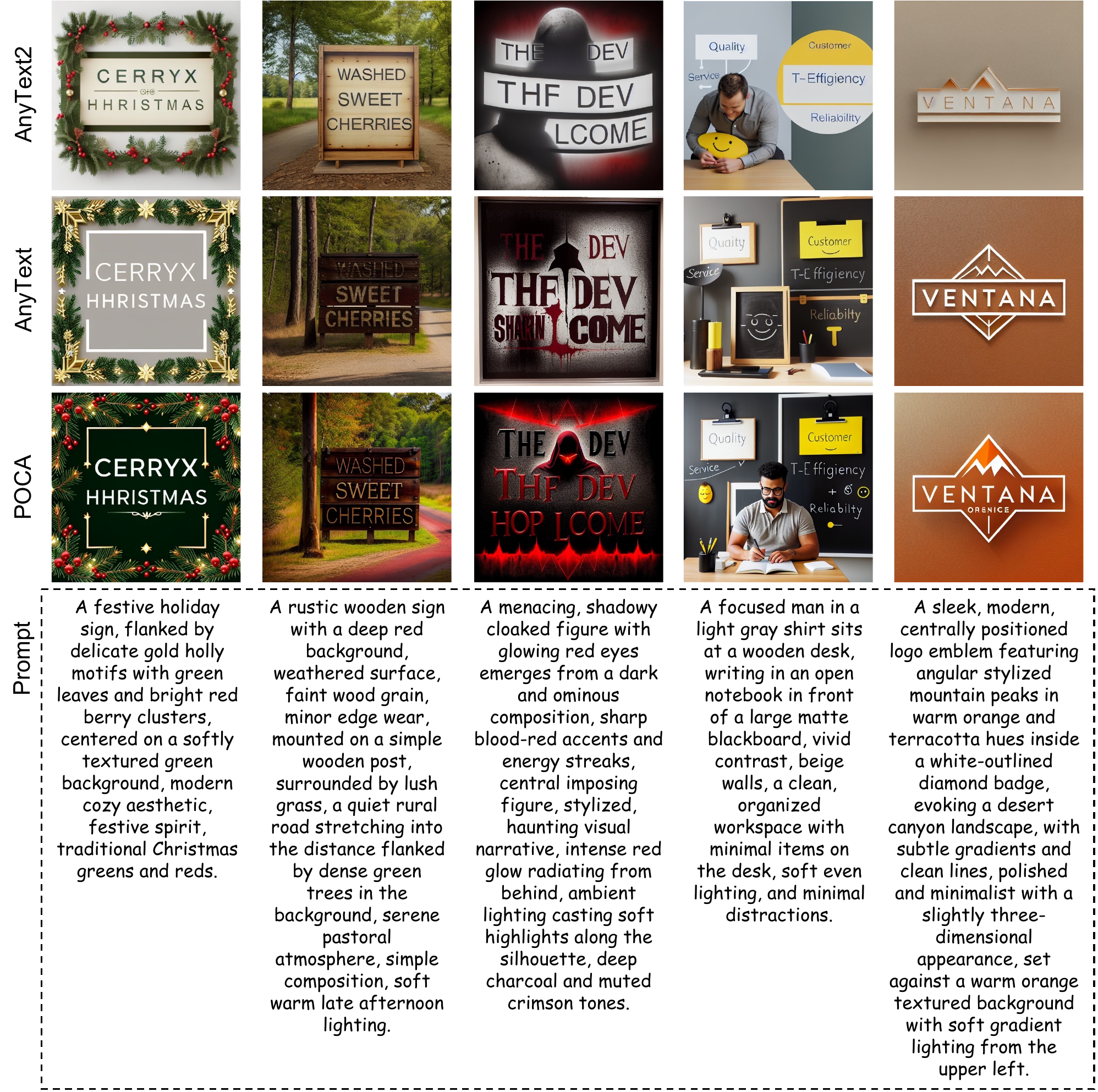}
  \caption{\textbf{Evaluation of complex instruction-following capability.} POCA demonstrates fidelity to instructions and higher control over fine-grained details under complex prompts, outperforming baselines on these challenging tasks.}
  \label{fig:supp_prompt_compare}
\end{figure*}

Finally, we use longer and more complex prompts in Fig.~\ref{fig:supp_prompt_compare} to test POCA, AnyText, and AnyText2 for their instruction-following capability. The results show that POCA generates the images most faithful to the instructions, such as the "gold holly motifs and red berry clusters" (column one), the "shadowy cloaked figure" (column three), and the "gradient lighting from the upper left" (column five). This large number of visual examples demonstrates that the images generated from POCA are improved across multiple reward criteria.

